\documentclass[11pt]{article}

\usepackage[final]{acl}

\usepackage{times}
\usepackage{latexsym}
\usepackage{amsmath}
\usepackage{booktabs}
\usepackage{multirow}
\usepackage[T1]{fontenc}

\usepackage[utf8]{inputenc}

\usepackage{microtype}

\usepackage{inconsolata}

\usepackage{graphicx}
\usepackage{enumitem}

\usepackage{tcolorbox}
\usepackage{xcolor}
\usepackage{amssymb}
\usepackage{algorithm}
\usepackage{algorithmic}

\title{GAM: Hierarchical Graph-based Agentic Memory for LLM Agents}

\author{
\textbf{Zhaofen Wu}$^{1}$\thanks{Equal contribution.} \quad
\textbf{Hanrong Zhang}$^{2}$\footnotemark[1]\textsuperscript{\hspace{0.2em}}\thanks{Corresponding authors.} \quad
\textbf{Fulin Lin}$^{1}$ \quad
\textbf{Wujiang Xu}$^{3}$ \quad
\textbf{Xinran Xu}$^{1}$\\
\textbf{Yankai Chen}$^{4,5}$ \quad
\textbf{Henry Peng Zou}$^{2}$ \quad
\textbf{Shaowen Chen}$^{1}$ \quad
\textbf{Weizhi Zhang}$^{2}$ \quad
\textbf{Xue Liu}$^{4,5}$\\
\textbf{Philip S. Yu}$^{2}$ \quad
\textbf{Hongwei Wang}$^{1}$\footnotemark[2]\\
$^{1}$Zhejiang University \quad
$^{2}$University of Illinois Chicago \quad
$^{3}$Rutgers University\\
$^{4}$MBZUAI \quad
$^{5}$McGill University\\
\texttt{\{zhaofen1.25, hongweiwang\}@intl.zju.edu.cn, \{hzhan135, psyu\}@uic.edu}
}

\begin{document}
\maketitle
\begin{abstract}
To sustain coherent long-term interactions, Large Language Model (LLM) agents must navigate the tension between acquiring new information and retaining prior knowledge.
Current unified stream-based memory systems facilitate context updates but remain vulnerable to interference from transient noise. 
Conversely, discrete structured memory architectures provide robust knowledge retention but often struggle to adapt to evolving narratives.
To address this, we propose \textbf{\textsc{GAM}}, a hierarchical \textbf{G}raph-based \textbf{A}gentic \textbf{M}emory framework that explicitly decouples memory encoding from consolidation to effectively resolve the conflict between rapid context perception and stable knowledge retention.
By isolating ongoing dialogue in an event progression graph and integrating it into a topic associative network only upon semantic shifts, our approach minimizes interference while preserving long-term consistency. Additionally, we introduce a graph-guided, multi-factor retrieval strategy to enhance context precision. Experiments on LoCoMo and LongDialQA indicate that our method consistently outperforms state-of-the-art baselines in both reasoning accuracy and efficiency.
\end{abstract}

\section{Introduction}
Large Language Model (LLM) agents have demonstrated remarkable capabilities in natural language understanding, long-horizon complex reasoning, and question answering~\cite{didolkar2024metacognitive,anwar2025remembr,ma2025large,zhang2026evoskillsselfevolvingagentskills,zhang2026expseek,zhang2025websearchagenticdeep}.
However, enabling these agents to maintain coherent long-term interactions remains a significant challenge~\cite{du2025context, yang2025continual,zou2026userschangemindevaluating,zou2025llmbasedhumanagentcollaborationinteraction}. A robust agentic memory must rapidly capture real-time interactions while safeguarding established knowledge from transient noise corruption~\cite{tan2025prospect,su2026dialoguetimetemporalsemantic}.
\begin{figure}[!b]
    \centering
    \includegraphics[width=\columnwidth]{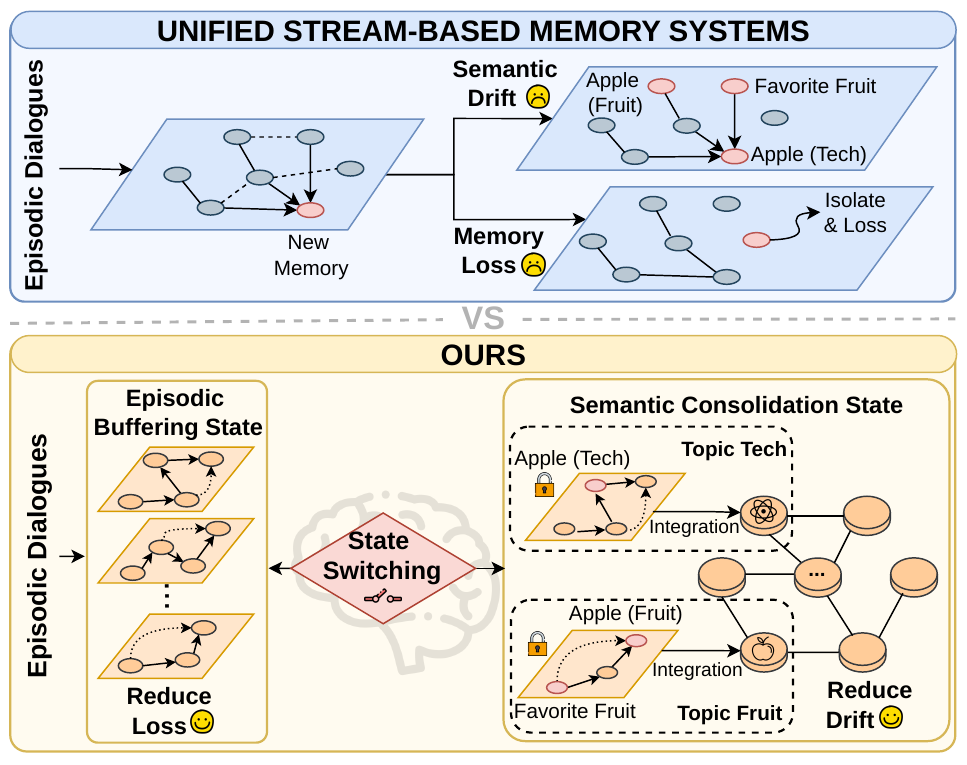} 
    \caption{Comparison between Unified Stream-based Memory Systems (upper) and GAM (lower). Unified architectures suffer from Memory Loss and Semantic Drift under direct updates, while GAM uses state switching to reduce both.}
    \label{fig:intro_comparison}
\end{figure}

Existing architectures often struggle to resolve this conflict. Recent approaches typically maintain a continuous, linear flow of information to handle ongoing interactions~\cite{zhong2024memorybank,lee2024human,packer2024memgptllmsoperatingsystems,kang2025memory}. We categorize these paradigms as Unified Stream-based Memory Systems. As depicted in the upper part of Figure \ref{fig:intro_comparison}, these methods perform updates directly on the active memory stream. This continuous exposure introduces a high risk of Memory Contamination~\cite{chen2025halumem}. 
Specifically, this continuous exposure leads to Memory Loss~\cite{jia2025evaluating}. In this phenomenon, established nodes become structurally isolated and forgotten due to weak connectivity. Furthermore, it causes Semantic Drift~\cite{yang2025drunkagent}. This occurs when distinct topics are erroneously conflated, thereby distorting the agent's thematic consistency.
Conversely, Discrete Structured Memory Architectures ensure stability by organizing information into rigid schemas~\cite{zhang2025g,edge2024local}. However, they often lack the agility to update narrative flows in real-time. This rigidity results in fragmented discourse representations. Consequently, they fail to capture the continuous evolution of long-term agent-user dialogues. 
While some recent studies attempt to address structural dynamism, they often rely on artificial discrete state changes or focus solely on retrieval optimization~\cite{anokhin2024arigraph,rasmussen2025zep}. These methods fail to fundamentally resolve the conflict between immediate encoding and long-term consolidation. In contrast, our approach explicitly decouples these processes to ensure robustness.

To address these challenges, we propose the Hierarchical Graph-based Agentic Memory (GAM) Framework which utilizes a Semantic-Event-Triggered mechanism to decouple encoding from consolidation. Drawing inspiration from sleep-dependent memory consolidation~\cite{chang2025sleep}, our architecture separates the agent memory lifecycle into two phases as illustrated in the lower part of Figure \ref{fig:intro_comparison}. An Episodic Buffering Phase constructs a local graph to capture real-time dependencies and strictly isolate transient context, while an event-driven Semantic Consolidation Phase integrates these details into the global network only upon semantic shifts. This ensures the agent global memory is updated solely with semantically complete units unlike prior stream-based methods. Finally, a Graph-Guided Multi-Factor Retrieval paradigm leverages structured cross-layer links to empower the agent to recall precise context.
The main contributions are as follows:
\begin{itemize}
    \item We propose the \textbf{GAM} framework, underpinned by a \textbf{Hierarchical Graph Memory Architecture}, which structurally mitigates the conflict between rapid context perception and stable knowledge retention. By physically separating the global Topic Associative Network from local Event Progression Graphs, we ensure robust write isolation for narrative buffering.
    
    \item We design a \textbf{State-Based Memory Consolidation} mechanism that dynamically transitions the system between Episodic Buffering State and Semantic Consolidation State. This strategy replaces arbitrary triggers with semantic divergence detection, ensuring that memory updates occur only at semantically complete boundaries to minimize contamination.
    
    \item We introduce a \textbf{Graph-Guided Multi-Factor Retrieval} strategy that effectively bridges the decoupled storage layers. By integrating temporal, confidence, and role-based signals into a top-down traversal, this mechanism allows the agent to recall precise episodic details grounded in high-level semantic themes.
    
    \item We demonstrate through extensive evaluations on LoCoMo and LongDialQA benchmarks that our approach outperforms state-of-the-art baselines in both reasoning accuracy and computational efficiency.
\end{itemize}

\begin{figure*}[t]
    \centering
    \includegraphics[width=\textwidth]{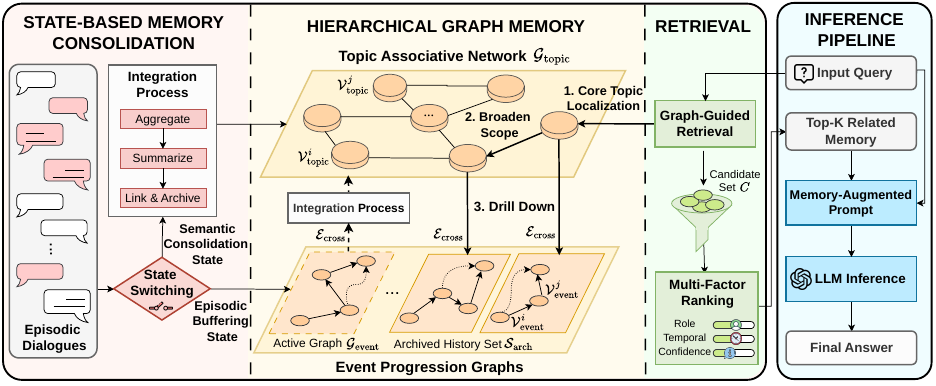}
    \caption{Architecture of GAM. The leftmost module switches between buffering and consolidation based on topic boundaries. The central Hierarchical Graph Memory decouples storage into a global Topic Associative Network and local Event Progression Graphs. The rightmost Graph-Guided Retrieval performs top-down traversal with multi-factor ranking.}
    \label{fig:architecture}
\end{figure*}
\section{Related Work}

\subsection{Unified Stream-based Memory}
Current LLM-based agents adopt a unified stream paradigm to maximize interaction plasticity~\cite{hu2026memoryageaiagents,huang2026rethinkingmemorymechanismsfoundation}. Early architectures like Generative Agents~\cite{park2023generative} record observations in a linear stream, relying on retrieval to reconstruct context. To manage infinite context windows, systems such as MemGPT~\cite{packer2024memgptllmsoperatingsystems}, MemoryOS~\cite{kang2025memory}, and Mem0~\cite{chhikara2025mem0} introduce hierarchical tiers inspired by operating systems, swapping information between working context and external storage. While these methods excel at rapid information encoding, they lack write isolation. Newly acquired and noisy information is directly appended or merged into the long-term store. This open-gate policy makes them susceptible to Memory Contamination~\cite{chen2025halumem}, where continuous updates cause semantic drift~\cite{yang2025continual} or memory loss. Even advanced self-reflective agents like A-Mem~\cite{xu2025mem} trigger updates based on arbitrary token counts or time-steps rather than semantic completeness, failing to prevent the corruption of stable knowledge by transient dialogue states.

\subsection{Discrete Structured Memory Architectures}
In contrast, Discrete Structured Memory Architectures prioritize stability through rigid knowledge representations. Systems like GraphRAG~\cite{edge2024local} and StructRAG~\cite{jia2025structrag} construct static knowledge graphs to facilitate precise multi-hop reasoning~\cite{li2025towards}. Approaches such as LightRAG~\cite{guo2024lightrag} and G-Memory~\cite{zhang2025g} further enhance this by indexing dual-level graph structures for complex entity retrieval. However, these architectures often suffer from rigidity and latency. The expensive index construction process required by traditional knowledge graphs makes them poorly suited for the fluid narrative flow of open-domain dialogue~\cite{ma2025large}. Consequently, they often produce fragmented discourse representations that excel at static fact retrieval but struggle to capture the continuous evolution of user states or sub-topic nuances in real-time.

\subsection{Approaches for Structural Dynamism}
Recent research has attempted to bridge the gap between stability and plasticity by introducing dynamic elements to structured memory. AriGraph~\cite{anokhin2024arigraph} introduces episodic nodes to track changing states but is tailored for text-based games with discrete state transitions, which are rarely present in the ambiguous boundaries of natural conversation. Similarly, Zep~\cite{rasmussen2025zep} addresses dynamism by focusing on optimizing the retrieval infrastructure rather than resolving the cognitive conflict between encoding and consolidation. Unlike these approaches, which rely on artificial state definitions or purely retrieval-based optimizations, our framework explicitly decouples the memory lifecycle, ensuring that global memory is updated only through consolidated and semantically complete units.
\section{Methodology}

\subsection{Problem Formulation and Overview}
\label{sec:problem_formulation}

We formalize memory management as an inference-time online decision process that balances immediate accessibility against long-term interference. Let a dialogue sequence $D = \{u_1, \dots, u_t\}$ be maintained by a hierarchical memory graph $\mathcal{H}_t$. Rather than optimizing model parameters, GAM seeks a policy that minimizes the cumulative operational cost of updating memory while preserving retrieval quality. For brevity, let $C_{\text{enc}}^{(t)}$ denote the operational cost $C_{\text{enc}}(u_t, \mathcal{H}_t)$, and let $C_{\text{inter}}^{(t)}$ denote the expected interference cost $C_{\text{inter}}(\mathcal{G}_{\text{topic}}^{(t)}, \mathcal{G}_{\text{event}}^{(t)})$.
\begin{equation}
    \pi^* = \arg\min_{\pi} \mathbb{E} \left[ \sum_{t=0}^{T} \left( C_{\text{enc}}^{(t)} + C_{\text{inter}}^{(t)} \right) \right].
    \label{eq:operational_cost}
\end{equation}
In the concrete architecture introduced below, the local role is instantiated by the Event Progression Graph $\mathcal{G}_{\text{event}}$, while the long-term global role is instantiated by the Topic Associative Network $\mathcal{G}_{\text{topic}}$.

Because evaluating future interference exactly is intractable online, we approximate the policy in Eq. \ref{eq:operational_cost} through a semantic divergence indicator $b_t \in \{0, 1\}$ that signals when the accumulated local context has drifted far enough from the global state to justify consolidation:
\begin{equation}
    b_t = \mathbb{I}(\Delta(\mathcal{G}_{\text{event}}^{(t)}, \mathcal{G}_{\text{topic}}^{(t)}) > \epsilon),
    \label{eq:memory_contamination}
\end{equation}
where $b_t=1$ signifies a structural shift requiring consolidation. Our objective is therefore not to solve Eq. \ref{eq:operational_cost} exactly, but to use $b_t$ as a low-cost policy trigger that reduces the long-run interference term without sacrificing fast online updates.

To address this optimization conflict, we propose the GAM framework. This framework employs a state-based consolidation mechanism within a hierarchical graph architecture to structurally separate the memory life cycle. As illustrated in Figure~\ref{fig:architecture}, our system satisfies the aforementioned objectives through three coordinated stages: (1) Episodic Buffering Phase. To optimize perception, we isolate ongoing dialogue in a local Event Progression Graph $\mathcal{G}_{\text{event}}$ whose structure is defined in Section \ref{sec:architecture}. This shielding mechanism protects the global store from temporary noise while the system operates in the Episodic Buffering State described in Section \ref{sec:consolidation}. (2) Semantic Consolidation Phase. Upon detecting a semantic indicator $b_t$, the system triggers an atomic update to ensure stability. As detailed in Section \ref{sec:consolidation}, buffered data is summarized and merged into the global Topic Associative Network $\mathcal{G}_{\text{topic}}$ only when narrative units are semantically complete. (3) Graph-Guided Retrieval. To bridge these separated layers during inference, we employ the multi-factor traversal strategy elaborated in Section \ref{sec:retrieval}. This process integrates temporal signals and confidence scores to achieve precise context recovery from both local details and global themes. The detailed end-to-end execution flow of GAM is provided in Algorithm \ref{alg:gam_process} in Appendix \ref{app:implementation_details}.

\subsection{Hierarchical Graph Memory Architecture}
\label{sec:architecture}

To implement the structural separation described in Section \ref{sec:problem_formulation}, we define the time-varying memory structure $\mathcal{H}_t$ as a composite graph. This architecture separates the storage of consolidated knowledge from the temporary context buffer and the historical archives. We formally define it as:
\begin{equation}
    \mathcal{H}_t = \{ \mathcal{G}_{\text{topic}}^{(t)}, \mathcal{G}_{\text{event}}^{(t)}, \mathcal{S}_{\text{arch}}^{(t)}, \mathcal{E}_{\text{cross}}^{(t)} \},
    \label{eq:memory_state}
\end{equation}
where $\mathcal{S}_{\text{arch}}^{(t)}$ represents the set of archived event graphs that have been consolidated, while $\mathcal{G}_{\text{event}}^{(t)}$ denotes the active buffer. This decomposition makes the abstract roles from Equation \ref{eq:operational_cost} explicit: $\mathcal{G}_{\text{event}}$ realizes the transient local state used for rapid perception, $\mathcal{G}_{\text{topic}}$ realizes the stable global substrate, and $\mathcal{S}_{\text{arch}}$ with $\mathcal{E}_{\text{cross}}$ preserve grounded evidence for later retrieval. We omit the time subscript $t$ when describing static structures for brevity.

The consolidated knowledge is represented as the Topic Associative Network which serves as the foundation for long-term retention. Its structure is defined as:
\begin{equation}
\mathcal{G}_{\text{topic}} = (\mathcal{V}_{\text{topic}}, \mathcal{E}_{\text{topic}}).
\end{equation}

In this layer, nodes $\mathcal{V}_{\text{topic}}$ represent high-level semantic clusters or abstract themes derived from historical interactions. The edges $\mathcal{E}_{\text{topic}}$ capture the deep semantic correlations between them. To capture subtle dependencies that heuristic metrics might miss, we quantify these edge weights via a semantic scorer based on large language models. This process utilizes the prompt detailed in Appendix \ref{app:prompt_relation}. While computationally intensive, this precise calculation is executed exclusively during the consolidation phase. This timing ensures that the structural precision of $\mathcal{G}_{\text{topic}}$ is achieved without compromising the interaction capabilities of the system.

In parallel, the temporary context is maintained in the active Event Progression Graph which is optimized for rapid updates. This structure is defined as follows:
\begin{equation}
\mathcal{G}_{\text{event}} = (\mathcal{V}_{\text{event}}, \mathcal{E}_{\text{event}}),
\end{equation}
where nodes $\mathcal{V}_{\text{event}}$ correspond to atomic interaction units such as user utterances or system responses which are captured in real-time. The edges $\mathcal{E}_{\text{event}}$ model the temporal and causal evolution of the current dialogue flow. Unlike the global layer, this structure is designed for rapid append operations which minimizes the encoding latency $C_{\text{enc}}$.

Finally, we establish cross-layer associations $\mathcal{E}_{\text{cross}}$ to bridge the gap between episodic details and semantic themes. These associations act as an index that connects topic nodes in the global graph to the archived snapshots of event graphs within $\mathcal{S}_{\text{arch}}$. This structure allows the retrieval mechanism to access the exact historical evidence associated with a theme without relying on the volatile active buffer.

\subsection{State-Based Memory Consolidation}
\label{sec:consolidation}

Our system models narrative dynamism as a finite-state machine that transitions between two distinct states based on semantic progression.

\paragraph{Implementation of Semantic Boundary Detection.}
To realize the theoretical boundary condition defined in Eq. \ref{eq:memory_contamination}, we require a mechanism to detect precisely when the semantic divergence exceeds the tolerance $\epsilon$. Since explicitly computing the high-dimensional distance $\Delta$ is computationally expensive and sensitive to lexical noise, we implement the detection function as a semantic discrimination task driven by an LLM $\mathcal{M}_{\theta}$.

In this framework, the LLM serves as a neural proxy for the thresholding function. The system maintains a fixed-capacity episodic buffer $B_t$ with a limit of 2048 tokens, derived from the linearized content of the current Event Progression Graph $\mathcal{G}_{\text{event}}$. Crucially, the discriminator is not queried at every turn. Instead, it is triggered only by sparse maintenance events such as session-end markers, natural interaction pauses, or buffer overflow. This process uses the structural instruction prompt defined in Appendix \ref{app:prompt_boundary}. In practice, $\epsilon$ is therefore not instantiated as a rigid numeric graph-distance threshold; the instruction strictness of the semantic discrimination prompt acts as a heuristic proxy for $\epsilon$. A positive detection where $b_t$ equals 1 serves as an indicator that the semantic distance has breached the stability threshold and triggers a transition to the Semantic Consolidation State. If no sparse trigger is observed before the buffer reaches capacity, overflow forces a consolidation check, and semantically related split segments can later be reconnected through strong topic-level edges.

\paragraph{Episodic Buffering State.}
In this state, the system focuses on capturing details of the ongoing dialogue. Incoming utterances are constructed as atomic event units $e_t$ and appended to the local graph by updating its node and edge sets:
\begin{equation}
\begin{aligned}
\mathcal{V}_{\text{event}}^{(t)}
&\leftarrow \mathcal{V}_{\text{event}}^{(t-1)} \cup \{e_t\},\\
\mathcal{E}_{\text{event}}^{(t)}
&\leftarrow \mathcal{E}_{\text{event}}^{(t-1)} \cup \mathcal{E}_{\text{temp}}^{(t)}.
\end{aligned}
\label{eq:event_update}
\end{equation}
where $\mathcal{E}_{\text{temp}}^{(t)}$ denotes the sequential edges that link the new event to the immediate context. This state functions as a strict write isolation buffer. By confining high-frequency updates to this local scope, we protect the global Topic Associative Network $\mathcal{G}_{\text{topic}}$ from temporary noise. This mechanism effectively separates the acquisition of new information from the modification of long-term memory. This separation mitigates the contamination risk outlined in Eq. \ref{eq:memory_contamination}.

\paragraph{Semantic Consolidation State.}
Upon detecting a topic boundary where $b_t$ equals 1, the system transitions to this state to consolidate the completed narrative unit. This process executes a graph merging operation that transforms the buffered subgraph $\mathcal{G}_{\text{event}}$ into a consolidated semantic node $v_{\text{new}}$. To address the trade-off between abstract reasoning and detailed recall, we design $v_{\text{new}}$ with a dual-granularity representation:
\begin{equation}
    v_{\text{new}} = \{ c_{\text{sum}}, c_{\text{raw}} \},
    \label{eq:topic_node}
\end{equation}
where $c_{\text{sum}}$ is generated by prompting an LLM with the instruction detailed in Appendix \ref{app:prompt_summary} to summarize the semantic content of the event graph. This summary enables high-level thematic reasoning. Simultaneously, $c_{\text{raw}}$ is constructed by concatenating the raw textual content of all event nodes within the buffer. This component preserves fine-grained details and prevents information loss during summarization.

Subsequently, the system integrates $v_{\text{new}}$ into the global network $\mathcal{G}_{\text{topic}}$ by establishing semantic edges. To avoid expensive full-graph reasoning over every topic node, we adopt a coarse-to-fine candidate selection strategy. The summary $c_{\text{sum}}$ first retrieves the top-$5$ nearest topic nodes using vector similarity, and only this compact candidate set is passed to the LLM-based semantic scorer introduced in Section \ref{sec:architecture}. Utilizing the instruction prompt detailed in Appendix \ref{app:prompt_relation}, we pair the summary of $v_{\text{new}}$ with each candidate topic node to query the model for a specific relationship type and a confidence score. New edges $\mathcal{E}_{\text{new}}$ are established when this confidence score exceeds a semantic threshold $\tau$. This coarse-to-fine procedure keeps the logical precision of LLM-based relation modeling while avoiding $O(N)$ graph-wide scoring. The global topology is updated as:
\begin{equation}
\begin{aligned}
\mathcal{V}_{\text{topic}}^{(t)}
&\leftarrow \mathcal{V}_{\text{topic}}^{(t-1)} \cup \{v_{\text{new}}\},\\
\mathcal{E}_{\text{topic}}^{(t)}
&\leftarrow \mathcal{E}_{\text{topic}}^{(t-1)} \cup \mathcal{E}_{\text{new}}.
\end{aligned}
\label{eq:topic_update}
\end{equation}
Once integration is complete, the current event graph structure is archived into the set $\mathcal{S}_{\text{arch}}^{(t)}$ and permanently linked to $v_{\text{new}}$ as grounded evidence through the creation of cross-layer edges $\mathcal{E}_{\text{cross}}$. The system then resets the active local buffer $\mathcal{G}_{\text{event}}$ to an empty state and reverts to the Episodic Buffering State for the subsequent narrative unit.

\subsection{Graph-Guided Multi-Factor Retrieval}
\label{sec:retrieval}

While the dual-phase architecture explicitly separates storage to promote stability, effective interaction requires a unified view of the memory system. We propose a graph-guided retrieval mechanism that acts as a logical bridge to traverse the hierarchical structure. This approach aims to retrieve context that is both semantically deep and temporally precise by integrating signals from the separated storage layers.

Unlike flat vector retrieval, our process exploits the topological structure of $\mathcal{H}_t$ in a top-down, expand-and-drill manner. The retrieval process operates in three stages:
\paragraph{Semantic Anchoring and Expansion.}
First, the system identifies the most relevant semantic anchors $\mathcal{V}_{\text{top}}$ in the Topic Associative Network $\mathcal{G}_{\text{topic}}$ via vector similarity. To capture latent dependencies beyond direct lexical matching, we leverage the global graph structure to expand the search scope. Specifically, we include the first-order neighbors of the top-$k$ nodes, utilizing the semantic edges $\mathcal{E}_{\text{topic}}$ to bring in contextually related themes. The final set of semantic anchors $\mathcal{V}_{\text{anchor}}$ is defined as:
\begin{equation}
\begin{aligned}
\mathcal{V}_{\text{anchor}}
&= \mathcal{V}_{\text{top}} \cup
\Bigl\{ v \in \mathcal{V}_{\text{topic}} \;\Bigm|\; \exists u \in \mathcal{V}_{\text{top}}, \\
&\qquad (u, v) \in \mathcal{E}_{\text{topic}} \Bigr\}.
\end{aligned}
\label{eq:anchor_set}
\end{equation}
\paragraph{Structural Drill-Down.}
From these expanded anchors, the system traverses the cross-layer links $\mathcal{E}_{\text{cross}}$ to access specific archived Event Progression Graphs stored in the archive set $\mathcal{S}_{\text{arch}}$. These archived graphs constitute the raw source of the topic nodes. Formally, we define the candidate set $\mathcal{C}$ by aggregating all event nodes from the archival graphs linked to any node in the anchor set:
\begin{equation}
\begin{aligned}
\mathcal{C}
&= \bigcup_{u \in \mathcal{V}_{\text{anchor}}}
\Bigl\{ v \in \mathcal{V}(\mathcal{G}') \;\Bigm|\; \mathcal{G}' \in \mathcal{S}_{\text{arch}}, \\
&\qquad (u, \mathcal{G}') \in \mathcal{E}_{\text{cross}} \Bigr\},
\end{aligned}
\label{eq:candidate_set}
\end{equation}
where $\mathcal{V}(\mathcal{G}')$ denotes the node set of an archived graph $\mathcal{G}'$. This strategy ensures that the agent accesses precise episodic details strictly aligned with both the direct and latent macro-level themes required by the query.

\paragraph{Multi-Factor Re-ranking.}
To rank the diverse candidates in $\mathcal{C}$, we employ a re-ranking strategy that combines deep semantic reasoning with explicit contextual signals. Given a user query $q$, we first compute a base semantic probability $P_{\text{sem}}(v|q)$ for each memory unit $v$ using a cross-encoder model. This model captures subtle semantic dependencies that bi-encoders might miss. To further enforce consistency with query constraints, we apply a multiplicative signal modulation mechanism defined below:
\begin{equation}
    \text{Score}(v, q) = P_{\text{sem}}(v|q) \cdot \prod_{k \in \mathcal{K}} \beta_k^{\mathbb{I}_{k}(v, q)},
\label{eq:retrieval_score}
\end{equation}
where $\mathcal{K}=\{\text{time}, \text{conf}, \text{role}\}$ indexes the set of modulation factors. The base probability is modulated by boosting factors $\beta_k > 1$ which are activated by indicator functions $\mathbb{I}_k(v, q) \in \{0, 1\}$. Specifically, we incorporate three pragmatic constraints to refine the retrieval scope. First, a temporal factor $\beta_{\text{time}}$ activates when the memory contains temporal expressions relevant to a time-sensitive query which promotes chronological precision. Second, an intrinsic confidence factor $\beta_{\text{conf}}$ which prioritizes information that successfully passed the self-consistency verification during its encoding phase. Finally, to address the complexity of multi-party interactions, we introduce a role-centric contextualization factor $\beta_{\text{role}}$. This factor helps disentangle mixed narrative threads by verifying if the memory source aligns with the target interlocutors implied by the query.

This multiplicative formulation ensures that candidates that match keywords without semantic relevance are not falsely promoted, as their base semantic probability remains low. We conducted a sensitivity analysis on these boosting factors $\beta$ from 1.0 to 2.0 to evaluate system robustness. As detailed in Appendix \ref{app:sensitivity}, our framework maintains performance within a wide hyperparameter margin. This result suggests that the integration of structured priors contributes to retrieval quality alongside semantic matching.

\section{Experiments}

\subsection{Experimental Settings}
\label{sec:experimental_settings}

\paragraph{Datasets.} We evaluate on two benchmarks for long-term interaction. LoCoMo~\cite{maharana2024evaluating} features long open-domain dialogues across five reasoning categories. Following MemoryOS~\cite{kang2025memory} and Mem0~\cite{chhikara2025mem0}, we focus on the first four categories to assess stability. LongDialQA~\cite{kim2024dialsim}, adapted from TV scripts like The Big Bang Theory, Friends, and The Office, simulates multi-party interactions to strictly assess context retention under adversarial settings.

\paragraph{Baselines.} We compare diverse architectures across three representative paradigms. First, heuristic approaches like MemoryBank~\cite{zhong2024memorybank} and ReadAgent~\cite{lee2024human} rely on passive forgetting curves or gist compression. Second, we benchmark unified stream-based systems, including OS-inspired models such as MemGPT~\cite{packer2024memgptllmsoperatingsystems} and MemoryOS~\cite{kang2025memory}, alongside the industrial standard Mem0~\cite{chhikara2025mem0}. Finally, we include the self-evolving agent A-Mem~\cite{xu2025mem} to assess recent self-reflective update mechanisms. To position GAM against explicit world-model graph memory, we provide a separate comparison with AriGraph in Appendix \ref{app:arigraph}.

\paragraph{Metrics and Implementation.}
We employ F1 Score for entity capture and BLEU-1 for lexical accuracy. Implemented via PyTorch and HuggingFace, we leverage Ollama and LiteLLM for efficient inference. We evaluate performance across models of varying scales, including Llama-3.2-3B-Instruct, Qwen2.5-7B-Instruct, Qwen2.5-14B-Instruct and GPT-4o-mini. All backbone LLMs are used as off-the-shelf instruct-tuned models, and GAM itself is a training-free memory framework without any additional fine-tuning of backbone parameters. The architecture uses all-MiniLM-L6-v2 sentence embeddings for indexing and cross-encoder/ms-marco-MiniLM-L-6-v2 for re-ranking. We set the retrieval size $k$ to 10 based on the sensitivity analysis detailed in Appendix \ref{app:retrieval_size}. Modulation factors are set to $\beta_{\text{time}}=1.4$, $\beta_{\text{role}}=1.4$, and $\beta_{\text{conf}}=1.2$, chosen from the stable regions in Appendix \ref{app:sensitivity} rather than aggressive per-dataset tuning. In implementation, the role indicator is triggered by lightweight speaker matching, while temporal and confidence cues are inferred with lightweight prompts. All experiments were conducted on NVIDIA RTX 4090 GPUs.

\subsection{Main Results}

\paragraph{Performance on LoCoMo.}

\begin{table*}[t]
  \centering
  \setlength{\tabcolsep}{3.5pt}
  { \small
  \begin{tabular}{llcccccccccc}
    \toprule
    \multirow{2}{*}{\textbf{Model}} & \multirow{2}{*}{\textbf{Method}} & \multicolumn{2}{c}{\textbf{Multi-Hop}} & \multicolumn{2}{c}{\textbf{Temporal}} & \multicolumn{2}{c}{\textbf{Open-Domain}} & \multicolumn{2}{c}{\textbf{Single-Hop}} & \multicolumn{2}{c}{\textbf{Avg}} \\
    \cmidrule(lr){3-4} \cmidrule(lr){5-6} \cmidrule(lr){7-8} \cmidrule(lr){9-10} \cmidrule(lr){11-12}
     & & \textbf{F1} & \textbf{BLEU-1} & \textbf{F1} & \textbf{BLEU-1} & \textbf{F1} & \textbf{BLEU-1} & \textbf{F1} & \textbf{BLEU-1} & \textbf{F1} & \textbf{BLEU-1} \\
    \midrule
    \multirow{7}{*}{Llama 3.2-3B} & ReadAgent & 2.05 & 1.96 & 4.33 & 4.33 & 7.53 & 6.39 & 3.52 & 2.75 & 4.36 & 3.86 \\
     & Memorybank & 7.35 & 5.34 & 3.90 & 3.41 & 4.91 & 5.42 & 8.76 & 7.04 & 6.23 & 5.30 \\
     & Memgpt & 7.59 & 5.09 & 3.25 & 3.15 & 5.74 & 5.55 & 7.91 & 6.64 & 6.12 & 5.11 \\
     & A-Mem & 17.44 & 11.74 & 26.38 & 19.50 & 12.53 & 11.83 & 28.14 & 23.87 & 21.12 & 16.73 \\
     & MemoryOS & 21.03 & 13.75 & 19.38 & 13.73 & 11.46 & 8.58 & 28.25 & 20.03 & 20.03 & 14.02 \\
     & Mem0 & \underline{27.51} & \underline{18.51} & \underline{30.08} & \underline{20.73} & \textbf{21.33} & \underline{15.00} & \underline{41.52} & \underline{30.89} & \underline{30.11} & \underline{21.28} \\ \cmidrule(lr){2-12}
     & \textbf{Ours} & \textbf{28.74} & \textbf{22.35} & \textbf{46.12} & \textbf{37.24} & \underline{21.18} & \textbf{18.33} & \textbf{49.02} & \textbf{42.18} & \textbf{36.27} & \textbf{30.02} \\
    \midrule
    \multirow{7}{*}{Qwen 2.5-7B} & ReadAgent & 5.37 & 4.02 & 2.82 & 2.74 & 5.86 & 4.78 & 5.23 & 4.13 & 4.82 & 3.92 \\
     & Memorybank & 6.22 & 5.39 & 2.58 & 2.27 & 8.53 & 7.58 & 6.17 & 4.87 & 5.88 & 5.03 \\
     & Memgpt & 6.17 & 4.97 & 2.18 & 2.77 & 6.33 & 5.93 & 8.95 & 7.33 & 5.91 & 5.25 \\
     & A-Mem & 21.85 & 13.40 & 27.54 & 22.49 & 14.16 & 13.06 & 33.25 & 29.00 & 24.20 & 19.49 \\
     & MemoryOS & 29.37 & 21.40 & 28.96 & 20.69 & \underline{20.22} & \underline{16.12} & 36.89 & 30.81 & 28.86 & 22.25 \\
     & Mem0 & \underline{32.86} & \underline{23.99} & \underline{41.22} & \underline{32.22} & 18.24 & 15.63 & \underline{49.21} & \underline{42.83} & \underline{35.38} & \underline{28.67} \\ \cmidrule(lr){2-12}
     & \textbf{Ours} & \textbf{35.32} & \textbf{25.77} & \textbf{48.97} & \textbf{39.68} & \textbf{21.14} & \textbf{17.89} & \textbf{54.58} & \textbf{48.62} & \textbf{40.00} & \textbf{32.99} \\
    \midrule
    \multirow{7}{*}{Qwen 2.5-14B} & ReadAgent & 4.33 & 2.99 & 2.66 & 2.71 & 4.93 & 5.53 & 5.16 & 4.05 & 4.27 & 3.82 \\
     & Memorybank & 3.65 & 3.16 & 2.59 & 2.33 & 7.33 & 7.45 & 4.71 & 3.72 & 4.57 & 4.17 \\
     & Memgpt & 8.37 & 5.42 & 4.05 & 3.64 & 11.81 & 11.52 & 11.70 & 9.79 & 8.98 & 7.59 \\
     & A-Mem & 26.55 & 16.56 & 28.30 & 23.02 & 16.74 & 14.89 & 36.95 & 32.58 & 27.14 & 21.76 \\
     & MemoryOS & 27.46 & 19.14 & 23.38 & 16.80 & \underline{19.27} & \underline{14.96} & 26.31 & 21.92 & 24.11 & 18.20 \\
     & Mem0 & \underline{31.45} & \underline{24.59} & \underline{50.20} & \textbf{39.85} & 17.66 & 14.43 & \underline{49.85} & \underline{43.78} & \underline{37.29} & \underline{30.66} \\ \cmidrule(lr){2-12}
     & \textbf{Ours} & \textbf{33.32} & \textbf{25.38} & \textbf{50.25} & \underline{39.33} & \textbf{20.40} & \textbf{18.50} & \textbf{57.55} & \textbf{50.99} & \textbf{40.38} & \textbf{33.55} \\
    \midrule
    \multirow{7}{*}{GPT-4o-mini} & ReadAgent & 9.15 & 6.48 & 12.60 & 8.87 & 5.31 & 5.12 & 9.67 & 7.66 & 9.18 & 7.03 \\
     & Memorybank & 5.00 & 4.77 & 9.68 & 6.99 & 5.56 & 5.94 & 6.61 & 5.16 & 6.71 & 5.71 \\
     & Memgpt & 26.65 & 17.72 & 25.52 & 19.44 & 9.15 & 7.44 & 41.04 & 34.34 & 25.59 & 19.73 \\
     & A-Mem & 27.02 & 20.09 & 45.85 & 36.67 & 12.14 & 12.00 & 44.65 & 37.06 & 32.41 & 26.45 \\
     & MemoryOS & \textbf{35.27} & \textbf{25.22} & 41.15 & 30.76 & 20.02 & \underline{16.52} & 48.62 & 42.99 & 36.27 & 28.87 \\
     & Mem0 & 32.78 & 20.87 & \textbf{56.42} & \textbf{45.89} & \underline{20.45} & 15.82 & \underline{51.81} & \underline{43.23} & \underline{40.37} & \underline{31.45} \\ \cmidrule(lr){2-12}
     & \textbf{Ours} & \textbf{35.88} & \textbf{27.96} & \underline{51.96} & \underline{42.26} & \textbf{28.12} & \textbf{24.51} & \textbf{56.58} & \textbf{51.18} & \textbf{43.14} & \textbf{36.48} \\
    \bottomrule
  \end{tabular}
  }
  \caption{Performance comparison on the LoCoMo dataset. We report F1 and BLEU-1 scores across different LLM backbones. The best results are highlighted in bold and the second best are underlined.}
  \label{tab:main_results_locomo}
\end{table*}

Table \ref{tab:main_results_locomo} presents the comparative results across four LLM backbones where our framework achieves the best average performance in most settings. Our method is particularly effective in complex reasoning tasks. For instance, with the Qwen 2.5-7B backbone, we surpass the strong commercial baseline Mem0~\cite{chhikara2025mem0} by over 18\% in Temporal F1 score. Even on the larger GPT-4o-mini model, where baselines perform strongly, our approach maintains the highest Average F1 score of 43.14, which demonstrates superior global consistency. These gains suggest that our dual-phase architecture provides a more robust structural foundation for long-term memory than purely retrieval-optimized systems. We note two informative exceptions. First, Mem0 remains stronger on GPT-4o-mini for Temporal F1 (56.42 versus 51.96), suggesting that when the backbone already has strong temporal reasoning and large context utilization, direct retrieval of raw traces can preserve exact chronological cues that GAM may partially compress during consolidation. Second, Open-Domain is consistently the most difficult category for GAM because these questions are weakly localized and less constrained by temporal or role cues, making broad thematic coverage more important than the structured pruning that benefits focused reasoning tasks.

\paragraph{Performance on LongDialQA.}

\begin{table}[t]
  \centering
  \setlength{\tabcolsep}{6pt}
  { \small
  \begin{tabular}{llcc}
    \toprule
    \multirow{2}{*}{\textbf{Model}} & \multirow{2}{*}{\textbf{Method}} & \multicolumn{2}{c}{\textbf{Avg}} \\
    \cmidrule(lr){3-4}
     & & \textbf{F1} & \textbf{BLEU-1} \\
    \midrule
    \multirow{7}{*}{Llama 3.2-3B} & ReadAgent & 3.97 & 4.04 \\
     & Memorybank & 3.93 & 3.60 \\
     & Memgpt & 3.90 & 3.36 \\
     & A-Mem & 4.11 & 3.27 \\
     & MemoryOS & 5.85 & 4.46 \\
     & Mem0 & \underline{6.85} & \underline{5.75} \\
     & \textbf{Ours} & \textbf{7.36} & \textbf{6.85} \\
    \midrule
    \multirow{7}{*}{Qwen 2.5-7B} & ReadAgent & 6.41 & 6.35 \\
     & Memorybank & 7.22 & 7.05 \\
     & Memgpt & 4.94 & 4.43 \\
     & A-Mem & 5.49 & 5.79 \\
     & MemoryOS & 6.76 & 5.11 \\
     & Mem0 & \underline{10.27} & \underline{9.91} \\
     & \textbf{Ours} & \textbf{12.55} & \textbf{12.43} \\
    \midrule
    \multirow{7}{*}{Qwen 2.5-14B} & ReadAgent & 8.71 & 8.85 \\
     & Memorybank & 9.86 & 10.24 \\
     & Memgpt & 5.83 & 5.81 \\
     & A-Mem & 8.53 & 7.93 \\
     & MemoryOS & 9.07 & 8.39 \\
     & Mem0 & \underline{11.48} & \underline{10.94} \\
     & \textbf{Ours} & \textbf{11.86} & \textbf{11.66} \\
    \midrule
    \multirow{7}{*}{GPT-4o-mini} & ReadAgent & 4.50 & 3.16 \\
     & Memorybank & 5.95 & 6.06 \\
     & Memgpt & 3.87 & 3.28 \\
     & A-Mem & 6.49 & 5.97 \\
     & MemoryOS & 10.61 & 9.31 \\
     & Mem0 & \underline{10.90} & \underline{9.84} \\
     & \textbf{Ours} & \textbf{11.18} & \textbf{10.14} \\
    \bottomrule
  \end{tabular}
  }
  \caption{Performance comparison on the LongDialQA dataset. We report average F1 and BLEU-1 scores. The best results are highlighted in bold and the second best are underlined.}
  \label{tab:main_results_longdialqa_avg}
  \vspace{-10pt}
\end{table}

As shown in Table \ref{tab:main_results_longdialqa_avg}, GAM consistently outperforms all baselines in complex multi-party narratives. Most notably, on the smaller Qwen 2.5-7B model, it achieves an Average F1 of 12.55, surpassing MemoryOS\cite{kang2025memory} by 86\%. This suggests that our structured memory effectively compensates for the limited context capacity of smaller models. Further analysis in Appendix \ref{app:longdialqa_analysis} confirms this advantage holds across all sub-datasets, indicating that our Semantic-Event-Triggered mechanism successfully mitigates interference from frequent speaker switching.

\subsection{Ablation Study}
To validate the contribution of each component, we performed an ablation study on the LoCoMo dataset using Qwen2.5-7B by removing the Topic Associative Network (w/o TAN), Event Progression Graph (w/o EPG), Semantic-Event-Triggered State Switching Mechanism (w/o SSM), and Multi-Factor Retrieval (w/o MFR). 

The average results presented in Table \ref{tab:ablation_avg} show that our full method consistently outperforms all variants. Detailed performance breakdowns across all task categories are provided in Appendix \ref{app:ablation_details}. The w/o EPG variant suffers from the most substantial degradation, suggesting that the narrative structure is fundamental for chronological robustness. The drop in w/o SSM underscores the necessity of decoupling encoding from consolidation to prevent contamination. Finally, declines in w/o TAN and w/o MFR validate that high-level topic organization and multi-signal fusion are essential for cross-session reasoning and precise retrieval.

\begin{table}[t]
\small
\centering
\setlength{\tabcolsep}{12pt} 
\begin{tabular}{lcc}
    \toprule
    \multirow{2}{*}{\textbf{Method}} & \multicolumn{2}{c}{\textbf{Avg}} \\
    \cmidrule{2-3}
     & \textbf{F1} & \textbf{B-1} \\
    \midrule
    w/o TAN & 35.07 & 29.00 \\
    w/o MFR & 35.94 & 29.28 \\
    w/o SSM & 32.58 & 26.13 \\
    w/o EPG & 25.06 & 20.76 \\
    \midrule
    \textbf{Ours} & \textbf{40.00} & \textbf{32.99} \\
    \bottomrule
\end{tabular}
\caption{Ablation study on the LoCoMo dataset. We report average F1 and BLEU-1 scores for different variants.}
\label{tab:ablation_avg}
\vspace{-10pt}
\end{table}

\begin{table}[t]
\small
\centering
\begin{tabular}{lccc}
\toprule
\textbf{Method} & \textbf{Tokens/Q} & \textbf{Time (s)} & \textbf{F1} \\
\midrule
A-Mem & 4221.12 & 2.21 & 24.20 \\
MemoryOS & 3400.41 & 154.22 & 28.86 \\
Mem0 & 1533.94 & \textbf{0.51} & 35.38 \\
\textbf{Ours} & \textbf{1370.18} & 0.80 & \textbf{40.00} \\
\bottomrule
\end{tabular}
\caption{Efficiency comparison on the LoCoMo dataset. We report token consumption, latency, and F1 score.}
\label{tab:efficiency}
\end{table}

\subsection{Topic Partitioning Strategies Analysis}

\begin{figure}[!t]
    \centering
    \includegraphics[width=0.93\linewidth]{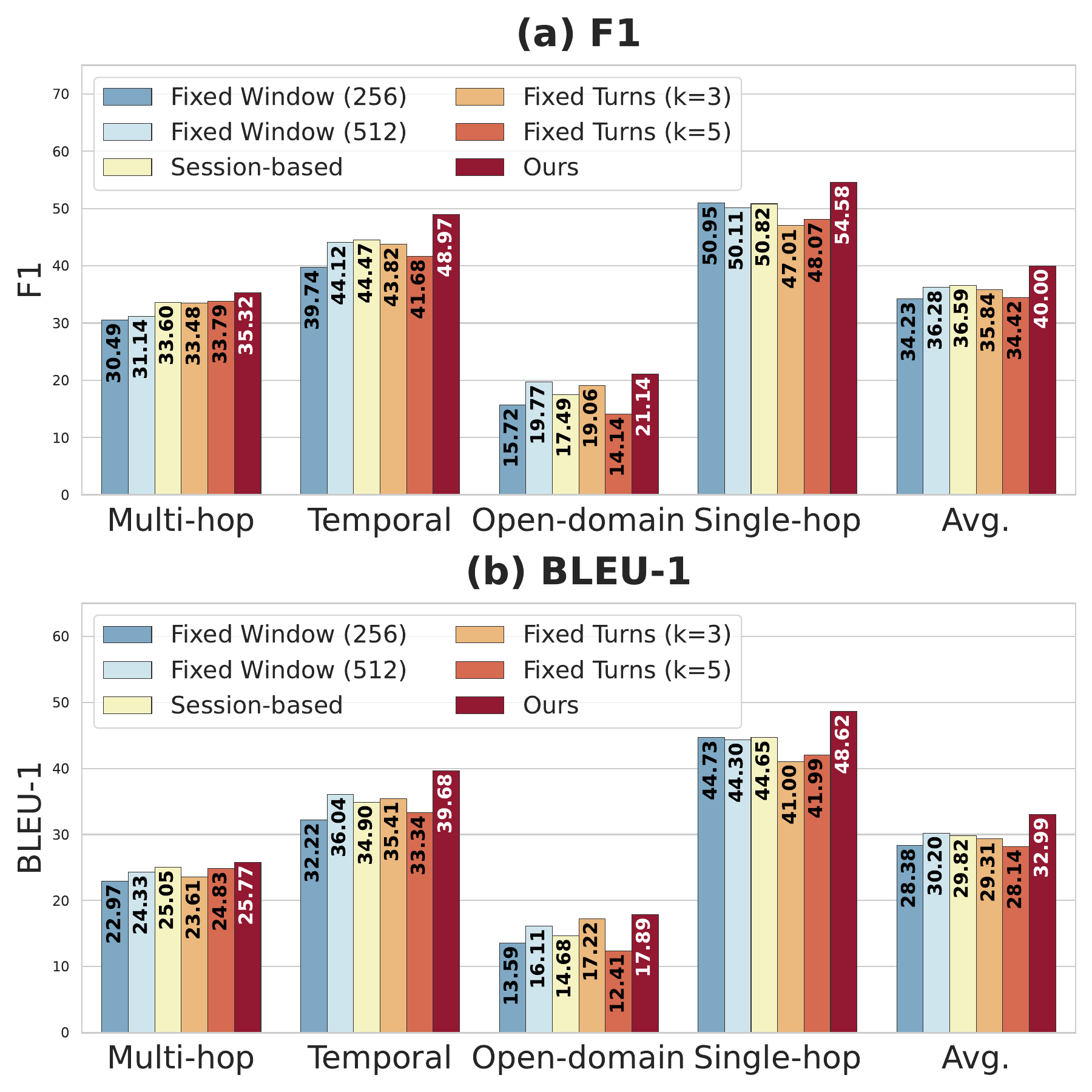}
    \caption{Impact of Partitioning Strategies on LoCoMo. Our semantic-boundary approach outperforms heuristic baselines across metrics.}
    \label{fig:topic_comparison}
    \vspace{-8pt}
\end{figure}

To validate the necessity of aligning memory consolidation with semantically complete boundaries, we compare our Semantic-Event-Triggered mechanism against three heuristic strategies: (1) \textbf{Fixed Window} (256 / 512 tokens), (2) \textbf{Fixed Turns} ($k \in \{3, 5\}$), and (3) \textbf{Session-based} partitioning. As shown in Figure \ref{fig:topic_comparison}, our method achieves the highest average F1 of 40.00, while Fixed Window (256) performs worst at 34.23. This confirms that arbitrary hard cuts fragment context and sever the Logical Flow edges essential for reasoning. The advantage is most pronounced on Temporal tasks, where our method improves over the best baseline from 44.12 to 48.97. Furthermore, it exceeds the strong Session-based baseline (36.59), showing that dynamic state switching captures finer semantic shifts within a single session, whereas coarse-grained session storage misses these nuances. Appendix \ref{app:noise} further shows that GAM remains stable and still outperforms fixed-window baselines under 40\% segmentation noise.

\subsection{Efficiency Analysis}
\label{sec:efficiency}
We evaluate the computational efficiency of different memory frameworks by analyzing the average token consumption and inference latency.
As shown in Table \ref{tab:efficiency}, our method achieves the lowest token consumption (1,370/query), reducing costs by 11\% compared to Mem0. While Mem0 shows slightly lower latency, our approach secures a superior trade-off by delivering a 13\% F1 gain at comparable speeds, while significantly outperforming A-Mem and MemoryOS in both metrics.

\section{Conclusion}
In this paper, we proposed GAM, a hierarchical graph-based agentic memory framework that mitigates the conflict between rapid context perception and stable knowledge retention. By decoupling narrative buffering from semantic consolidation through state-based switching, GAM updates memory only at semantically complete boundaries, isolating transient noise. Combined with graph-guided retrieval that fuses temporal, confidence, and role-centric signals, GAM achieves superior reasoning accuracy and efficiency over strong baselines. In the future, we plan to extend the current text-based memory into a native multimodal memory framework to support richer long-term reasoning over textual, visual, and auditory signals.

\section*{Limitations}
Although our experiments on LoCoMo and LongDialQA verify the effectiveness of the proposed architecture, the current instantiation of GAM is limited to text modality. This unimodal design limits the system's ability to leverage the rich nonverbal information commonly found in real-world interactions, such as visual context or acoustic cues. Consequently, the model cannot perform fine-grained reasoning over visual or auditory inputs. A native multimodal extension would replace purely textual memory representations with multimodal nodes that combine symbolic summaries with modality-specific features such as keyframe or audio representations. The same state-switching mechanism could then detect semantic shifts from visual or acoustic signals instead of only lexical topic changes. We leave this extension to future work and do not claim native multimodal reasoning in the current system.

\section*{Ethical Considerations}
Persistent memory systems for conversational agents raise meaningful privacy and safety concerns. First, users may reveal sensitive personal details that should not automatically enter long-term storage. GAM does not solve this issue by itself, but its state-based consolidation design creates a natural intervention point: before buffered content is written into the global topic graph, privacy filters or user-defined retention rules can inspect and block sensitive content.

Second, long-term memory systems should support user control over what is stored. Because GAM organizes memory as explicit graph nodes and cross-layer links rather than only opaque vector states, it offers a clearer substrate for inspection, selective deletion, freezing, or branch-level pruning. This does not guarantee that a deployed system will expose such controls, but it makes them architecturally easier to implement.

Third, memory-bearing agents may form inaccurate beliefs about users through faulty summarization or incorrect semantic links. In a graph-based memory, such failures can in principle be corrected locally by removing spurious edges, adjusting confidence weights, or deleting incorrect topic nodes, without retraining the entire backbone model. We view these affordances as useful safety primitives, not as complete solutions; robust deployment still requires careful interface design, auditing, and user consent mechanisms.

\section*{Acknowledgments}
This work was supported in part by the National Key Research and Development Program of China under Grants No. 2024YFF0907802 and 2024YFF0907803, and by the National Natural Science Foundation of China under Grant No. 62276230. It was also supported in part by the U.S. National Science Foundation (NSF) under Grants III-2106758 and POSE-2346158, as well as by the ACCESS program under allocation CIS260354.


\appendix
\section{Implementation Details}
\label{app:implementation_details}

\subsection{End-to-End Execution Flow}

For completeness, we summarize the online maintenance and inference workflow of GAM below.

\begin{algorithm}[ht]
\caption{The GAM Execution Flow: Memory Maintenance and Retrieval}
\label{alg:gam_process}
\begin{algorithmic}[1]
\REQUIRE Dialogue stream $D=\{u_1, \dots\}$, Query $q$, Thresholds $\epsilon, \tau$
\ENSURE Response to $q$, Updated Memory $\mathcal{H}_t$

\STATE \textbf{Process I: Memory Maintenance (Online)}
\FOR{each incoming utterance $u_t$ in $D$}
    \STATE Update the active event graph with $e_t$ by Eq. \ref{eq:event_update}
    \STATE Compute boundary indicator $b_t$ by Eq. \ref{eq:memory_contamination}
    \IF{$b_t = 1$}
        \STATE Generate topic node $v_{\text{new}}$ by Eq. \ref{eq:topic_node}
        \STATE Update the topic graph by Eq. \ref{eq:topic_update}
        \STATE Archive buffer: $\mathcal{S}_{\text{arch}}^{(t)} \leftarrow \mathcal{S}_{\text{arch}}^{(t-1)} \cup \{\mathcal{G}_{\text{event}}^{(t)}\}$
        \STATE Establish cross-layer index: $\mathcal{E}_{\text{cross}}^{(t)} \leftarrow \mathcal{E}_{\text{cross}}^{(t-1)} \cup \{(v_{\text{new}}, \mathcal{G}_{\text{event}}^{(t)})\}$
        \STATE Reset buffer: $\mathcal{G}_{\text{event}}^{(t)} \leftarrow \emptyset$
    \ENDIF
\ENDFOR

\STATE \textbf{Process II: Graph-Guided Retrieval (Inference)}
\STATE Identify top-$k$ anchors $\mathcal{V}_{\text{top}}$ from $\mathcal{G}_{\text{topic}}$ via similarity
\STATE Expand scope to semantic anchors $\mathcal{V}_{\text{anchor}}$ by Eq. \ref{eq:anchor_set}
\STATE Extract candidate set $\mathcal{C}$ by Eq. \ref{eq:candidate_set}
\STATE Rank candidates $v \in \mathcal{C}$ by Eq. \ref{eq:retrieval_score}
\RETURN Top-ranked context for generation
\end{algorithmic}
\end{algorithm}

\subsection{Event-Driven Boundary Detection}

To improve reproducibility, we summarize the operational trigger logic used by the semantic boundary detector. Rather than scanning every incoming turn, GAM maintains a bounded episodic buffer and invokes the discriminator only when a sparse maintenance event occurs.

\begin{algorithm}[!htbp]
\caption{Event-Driven Semantic Boundary Detection}
\label{alg:event_driven_detection}
\begin{algorithmic}[1]
\REQUIRE Incoming turn $u_t$, token limit $T_{\max}=2048$
\STATE Append $u_t$ to the local buffer $B_t$ and update the active event graph as in Eq. \ref{eq:event_update}
\IF{SessionEnd$(u_t)$ or TokenCount$(B_t) > T_{\max}$}
    \STATE Estimate the proxy boundary indicator $b_t$ for Eq. \ref{eq:memory_contamination} via $\text{LLM}_{\text{detect}}(\text{Prompt}_{\text{boundary}}, B_t)$
    \IF{$b_t = 1$}
        \STATE Trigger semantic consolidation for the buffered content
    \ENDIF
    \STATE Reset or trim the local buffer according to the consolidation result
\ENDIF
\end{algorithmic}
\end{algorithm}

\subsection{Coarse-to-Fine Topic Linking}

The consolidation stage does not score a new semantic node against every existing topic node. Instead, GAM narrows the candidate set through vector retrieval and then applies LLM-based relation scoring only to a small candidate pool.

\begin{algorithm}[!htbp]
\caption{Coarse-to-Fine Candidate Selection for Topic Linking}
\label{alg:coarse_to_fine_linking}
\begin{algorithmic}[1]
\REQUIRE New semantic node $v_{\text{new}}$, global graph $\mathcal{G}_{\text{topic}}$, retrieval depth $K=5$, threshold $\tau$
\STATE Retrieve top-$K$ candidates using the summary field $c_{\text{sum}}$ from Eq. \ref{eq:topic_node}
\STATE $C \leftarrow \text{VectorRetrieve}(\mathcal{G}_{\text{topic}}, c_{\text{sum}}, \text{top\_k}=K)$
\FOR{each topic node $v$ in $C$}
    \STATE $(w, r) \leftarrow \text{LLM}_{\text{score}}(\text{Prompt}_{\text{relation}}, v_{\text{new}}, v)$
    \IF{$w > \tau$}
        \STATE Add edge $(v_{\text{new}}, v)$ with relation $r$ and weight $w$
    \ENDIF
\ENDFOR
\STATE Update the topic graph by Eq. \ref{eq:topic_update}
\end{algorithmic}
\end{algorithm}

\section{Detailed Experimental Results}
\subsection{Detailed Analysis of LongDialQA Results}
\label{app:longdialqa_analysis}

\begin{table*}[t]

  \centering

  \setlength{\tabcolsep}{3.5pt}

  { \small

  \begin{tabular}{llcccccccc}

    \toprule

    \multirow{2}{*}{\textbf{Model}} & \multirow{2}{*}{\textbf{Method}} & \multicolumn{2}{c}{\textbf{Big Bang Theory}} & \multicolumn{2}{c}{\textbf{Friends}} & \multicolumn{2}{c}{\textbf{The Office}} & \multicolumn{2}{c}{\textbf{Avg}} \\

    \cmidrule(lr){3-4} \cmidrule(lr){5-6} \cmidrule(lr){7-8} \cmidrule(lr){9-10}

     & & \textbf{F1} & \textbf{B-1} & \textbf{F1} & \textbf{B-1} & \textbf{F1} & \textbf{B-1} & \textbf{F1} & \textbf{B-1} \\

    \midrule

    \multirow{7}{*}{Llama 3.2-3B} & ReadAgent & 5.19 & 4.70 & 3.49 & 4.24 & 3.24 & 3.18 & 3.97 & 4.04 \\

     & Memorybank & 4.77 & 4.46 & \underline{4.14} & 3.77 & 2.87 & 2.56 & 3.93 & 3.60 \\

     & Memgpt & 3.94 & 2.97 & 3.99 & 3.85 & 3.77 & 3.25 & 3.90 & 3.36 \\

     & A-Mem & 5.69 & 4.20 & 3.79 & 3.32 & 2.86 & 2.29 & 4.11 & 3.27 \\

     & MemoryOS & 8.65 & 6.46 & 2.98 & 3.56 & 5.92 & 3.35 & 5.85 & 4.46 \\

     & Mem0 & \textbf{10.84} & \underline{6.63} & 3.50 & \underline{5.37} & \underline{6.20} & \underline{5.25} & \underline{6.85} & \underline{5.75} \\

     & \textbf{Ours} & \underline{9.42} & \textbf{8.85} & \textbf{4.83} & \textbf{5.70} & \textbf{7.82} & \textbf{5.99} & \textbf{7.36} & \textbf{6.85} \\

    \midrule

    \multirow{7}{*}{Qwen 2.5-7B} & ReadAgent & 11.67 & 11.40 & 2.74 & 3.18 & 4.83 & 4.46 & 6.41 & 6.35 \\

     & Memorybank & 9.11 & 9.15 & \underline{7.29} & \underline{7.81} & 5.27 & 4.18 & 7.22 & 7.05 \\

     & Memgpt & 6.22 & 5.87 & 5.23 & 4.36 & 3.37 & 3.05 & 4.94 & 4.43 \\

     & A-Mem & 7.93 & 8.09 & 4.85 & 5.87 & 3.68 & 3.42 & 5.49 & 5.79 \\

     & MemoryOS & 9.71 & 8.48 & 5.71 & 3.20 & 4.86 & 3.64 & 6.76 & 5.11 \\

     & Mem0 & \underline{15.01} & \underline{14.97} & 6.69 & 6.14 & \underline{9.11} & \underline{8.63} & \underline{10.27} & \underline{9.91} \\

     & \textbf{Ours} & \textbf{18.73} & \textbf{18.35} & \textbf{7.32} & \textbf{8.22} & \textbf{11.60} & \textbf{10.72} & \textbf{12.55} & \textbf{12.43} \\

    \midrule

    \multirow{7}{*}{Qwen 2.5-14B} & ReadAgent & 13.85 & 13.79 & 7.11 & \underline{7.71} & 5.18 & 5.05 & 8.71 & 8.85 \\

     & Memorybank & 17.58 & 17.56 & 4.60 & 6.03 & 7.41 & 7.12 & 9.86 & 10.24 \\

     & Memgpt & 8.99 & 8.87 & 3.91 & 4.14 & 4.60 & 4.43 & 5.83 & 5.81 \\

     & A-Mem & 14.88 & 14.04 & 6.48 & 5.78 & 4.23 & 3.98 & 8.53 & 7.93 \\

     & MemoryOS & 16.36 & 15.70 & 6.38 & 5.91 & 4.48 & 3.56 & 9.07 & 8.39 \\

     & Mem0 & \textbf{18.32} & \underline{17.58} & \underline{7.22} & 7.68 & \underline{8.90} & \underline{7.57} & \underline{11.48} & \underline{10.94} \\

     & \textbf{Ours} & \underline{18.13} & \textbf{17.73} & \textbf{7.36} & \textbf{8.03} & \textbf{10.10} & \textbf{9.22} & \textbf{11.86} & \textbf{11.66} \\

    \midrule

    \multirow{7}{*}{GPT-4o-mini} & ReadAgent & 6.65 & 5.89 & 3.12 & 1.18 & 3.73 & 2.40 & 4.50 & 3.16 \\

     & Memorybank & 7.98 & 8.21 & 6.41 & 7.42 & 3.46 & 2.54 & 5.95 & 6.06 \\

     & Memgpt & 5.27 & 4.80 & 1.83 & 2.29 & 4.52 & 2.75 & 3.87 & 3.28 \\

     & A-Mem & 6.22 & 5.60 & 6.80 & 7.29 & 6.44 & 5.01 & 6.49 & 5.97 \\

     & MemoryOS & 13.36 & 11.63 & \underline{9.56} & 8.16 & 8.92 & 8.13 & 10.61 & 9.31 \\

     & Mem0 & \underline{13.73} & \underline{12.77} & 9.42 & \underline{8.28} & \textbf{9.54} & \textbf{8.46} & \underline{10.90} & \underline{9.84} \\

     & \textbf{Ours} & \textbf{14.70} & \textbf{13.70} & \textbf{9.64} & \textbf{8.34} & \underline{9.21} & \underline{8.39} & \textbf{11.18} & \textbf{10.14} \\

    \bottomrule

  \end{tabular}

  }

  \caption{Detailed performance comparison on the LongDialQA dataset, including the newly added Mem0 baseline. We report F1 and BLEU-1 scores on three sub-datasets (Big Bang Theory, Friends, The Office) and their average. The best results for each backbone are highlighted in bold, and the second best are underlined.}

  \label{tab:appendix_results_longdialqa}

\end{table*}

Table \ref{tab:appendix_results_longdialqa} presents the comprehensive performance breakdown on the LongDialQA benchmark across three distinct sub-datasets: The Big Bang Theory, Friends, and The Office. These datasets represent highly complex multi-party environments with frequent speaker switching and interleaved plotlines.

Our framework demonstrates consistent superiority over state-of-the-art baselines across most configurations. Specifically, on the Friends and The Office sub-datasets, which are characterized by nuanced interpersonal dynamics and long-term character arcs, our method achieves significantly higher F1 scores compared to the strong commercial baseline Mem0. For instance, using the Qwen 2.5-7B backbone, our approach surpasses Mem0 by a notable margin on the The Office subset (11.60 vs 9.11). This indicates that the proposed Role-Centric Contextualization factor effectively disentangles narrative threads in dense multi-speaker scenarios, preventing the confusion often observed in OS-based or heuristic memory systems.

Furthermore, the results highlight the model-agnostic robustness of our architecture. While baselines like A-Mem and MemoryOS exhibit performance fluctuations when scaling down from GPT-4o-mini to Llama 3.2-3B, our framework maintains a stable performance advantage. Even on the smaller Llama 3.2-3B model, our method achieves the highest Average F1 score (7.36), suggesting that the structured graph memory provides a crucial external knowledge scaffold that compensates for the limited internal context capacity of smaller LLMs.

\subsection{Detailed Ablation Study Results}
\label{app:ablation_details}

Table \ref{tab:ablation_full} presents the comprehensive breakdown of our ablation study on the LoCoMo dataset. We evaluate the contribution of each component across four specific reasoning categories: Multi-Hop, Temporal, Open-Domain, and Single-Hop.

\begin{table*}[htbp]
\small
  \centering
  \begin{tabular}{lcccccccccc}
    \toprule
    \multirow{2}{*}{\textbf{Method}} & \multicolumn{2}{c}{\textbf{Multi-Hop}} & \multicolumn{2}{c}{\textbf{Temporal}} & \multicolumn{2}{c}{\textbf{Open-Domain}} & \multicolumn{2}{c}{\textbf{Single-Hop}} & \multicolumn{2}{c}{\textbf{Avg}} \\
    \cmidrule{2-11}
     & \textbf{F1} & \textbf{B-1} & \textbf{F1} & \textbf{B-1} & \textbf{F1} & \textbf{B-1} & \textbf{F1} & \textbf{B-1} & \textbf{F1} & \textbf{B-1} \\
    \midrule
    w/o TAN & 31.80 & 21.29 & 39.74 & 34.15 & 16.28 & 13.42 & 52.46 & 47.12 & 35.07 & 29.00 \\
    w/o MFR & 33.73 & 22.91 & 46.12 & 38.06 & 16.30 & 14.20 & 47.61 & 41.96 & 35.94 & 29.28 \\
    w/o SSM & 30.32 & 19.96 & 39.53 & 32.55 & 16.73 & 14.08 & 43.72 & 37.93 & 32.58 & 26.13 \\
    w/o EPG & 25.35 & 18.94 & 33.61 & 27.70 & 14.59 & 13.44 & 26.68 & 22.97 & 25.06 & 20.76 \\
    \midrule
    \textbf{Ours} & \textbf{35.32} & \textbf{25.77} & \textbf{48.97} & \textbf{39.68} & \textbf{21.14} & \textbf{17.89} & \textbf{54.58} & \textbf{48.62} & \textbf{40.00} & \textbf{32.99} \\
    \bottomrule
  \end{tabular}
    \caption{Detailed ablation study results across different task categories on the LoCoMo dataset.}
  \label{tab:ablation_full}
\end{table*}

\subsection{Comparison with AriGraph on LoCoMo}
\label{app:arigraph}

To further position GAM against graph-based memory approaches with explicit world modeling, we implement AriGraph as an additional baseline on LoCoMo. While AriGraph is effective when state transitions are externally identifiable, open-domain dialogue presents fuzzy topic boundaries and temporally entangled evidence, making semantic consolidation substantially more challenging.

\begin{table*}[!htbp]
\caption{Performance comparison with AriGraph on the LoCoMo benchmark. We report F1 and BLEU-1 for each reasoning category and their average.}
\small
\centering
\setlength{\tabcolsep}{3.5pt}
\begin{tabular}{llcccccccccc}
\toprule
\multirow{2}{*}{\textbf{Model}} & \multirow{2}{*}{\textbf{Method}} & \multicolumn{2}{c}{\textbf{Multi-Hop}} & \multicolumn{2}{c}{\textbf{Temporal}} & \multicolumn{2}{c}{\textbf{Open-Domain}} & \multicolumn{2}{c}{\textbf{Single-Hop}} & \multicolumn{2}{c}{\textbf{Avg}} \\
\cmidrule(lr){3-4} \cmidrule(lr){5-6} \cmidrule(lr){7-8} \cmidrule(lr){9-10} \cmidrule(lr){11-12}
 & & \textbf{F1} & \textbf{B-1} & \textbf{F1} & \textbf{B-1} & \textbf{F1} & \textbf{B-1} & \textbf{F1} & \textbf{B-1} & \textbf{F1} & \textbf{B-1} \\
\midrule
\multirow{2}{*}{Llama 3.2-3B} & AriGraph & 13.99 & 7.83 & 3.72 & 2.69 & 20.10 & 17.74 & 14.63 & 11.67 & 12.58 & 9.47 \\
 & \textbf{Ours} & \textbf{28.74} & \textbf{22.35} & \textbf{46.12} & \textbf{37.24} & \textbf{21.18} & \textbf{18.33} & \textbf{49.02} & \textbf{42.18} & \textbf{36.27} & \textbf{30.02} \\
\midrule
\multirow{2}{*}{Qwen 2.5-7B} & AriGraph & 21.32 & 12.30 & 5.51 & 5.25 & 15.10 & 12.40 & 24.44 & 19.49 & 19.34 & 14.77 \\
 & \textbf{Ours} & \textbf{35.32} & \textbf{25.77} & \textbf{48.97} & \textbf{39.68} & \textbf{21.14} & \textbf{17.89} & \textbf{54.58} & \textbf{48.62} & \textbf{40.00} & \textbf{32.99} \\
\midrule
\multirow{2}{*}{Qwen 2.5-14B} & AriGraph & 29.40 & 18.20 & 6.00 & 5.01 & 16.87 & 13.65 & 30.68 & 24.38 & 24.44 & 18.54 \\
 & \textbf{Ours} & \textbf{33.32} & \textbf{25.38} & \textbf{50.25} & \textbf{39.33} & \textbf{20.40} & \textbf{18.50} & \textbf{57.55} & \textbf{50.99} & \textbf{40.38} & \textbf{33.55} \\
\midrule
\multirow{2}{*}{GPT-4o-mini} & AriGraph & 25.10 & 12.92 & 8.44 & 6.00 & 15.32 & 11.53 & 30.93 & 24.52 & 24.20 & 17.72 \\
 & \textbf{Ours} & \textbf{35.88} & \textbf{27.96} & \textbf{51.96} & \textbf{42.26} & \textbf{28.12} & \textbf{24.51} & \textbf{56.58} & \textbf{51.18} & \textbf{43.14} & \textbf{36.48} \\
\bottomrule
\end{tabular}
\label{tab:arigraph_locomo}
\end{table*}

Across all backbones, GAM consistently surpasses AriGraph, with the largest gap appearing in Temporal reasoning. This supports the claim that a semantic-boundary-triggered memory lifecycle is better suited to continuous dialogue streams than graph updates designed for explicitly observable discrete states.

\subsection{MovieChat-1K Proxy Multimodal Experiment}
\label{app:moviechat}

To probe the modality-agnostic structure of GAM without redesigning the full system into a native multimodal model, we conducted a proxy experiment on MovieChat-1K using dense video captions as a continuous visual narrative stream. We compare GAM against a long-context direct-input baseline under the Qwen2.5-7B backbone with retrieval depth $K=15$. The same temporal, role, and confidence retrieval factors used for text were transferred in a zero-shot manner by aligning them with timestamps, on-screen characters, and action coherence.

\begin{table}[!htbp]
\caption{Proxy multimodal evaluation on MovieChat-1K using dense video captions as the input stream.}
\small
\centering
\begin{tabular}{lccc}
\toprule
\textbf{Method} & \textbf{Accuracy} & \textbf{F1} & \textbf{B-1} \\
\midrule
Baseline (Direct Input) & 47.35 & 42.35 & 45.31 \\
\textbf{Ours (GAM)} & \textbf{55.51} & \textbf{48.95} & \textbf{51.63} \\
Improvement & +8.16 & +6.60 & +6.32 \\
\bottomrule
\end{tabular}
\label{tab:moviechat}
\end{table}

The proxy results suggest that the hierarchical graph memory and state-based consolidation mechanism are not tied to textual dialogue alone. Even without domain-specific tuning, GAM improves both accuracy and F1 on long-form visual narratives represented as caption streams, supporting the multimodal roadmap discussed in the limitations section.

\section{System Analysis}
\label{app:system_analysis}

\subsection{Operational Cost of Boundary Detection}
\label{app:detection_cost}

The efficiency table in the main text reports query-time inference cost. To complement it, we separately measure the maintenance-stage cost of the semantic boundary detector on LoCoMo. Because the detector is triggered sparsely rather than at every turn, the additional overhead remains small relative to the main inference pipeline.

\begin{table}[!htbp]
\caption{Cost analysis of the semantic divergence detection mechanism on LoCoMo.}
\small
\centering
\begin{tabular}{lc}
\toprule
\textbf{Metric} & \textbf{Value} \\
\midrule
Avg. Sessions per Sample & 27.2 \\
Avg. Latency per Session & 0.63 s \\
Avg. Input Tokens per Session & 932.36 \\
Avg. Output Tokens per Session & 6.15 \\
\bottomrule
\end{tabular}
\label{tab:detection_cost}
\end{table}

These results confirm that the detector behaves as a sparse maintenance primitive rather than a high-frequency online bottleneck. Combined with the event-driven trigger logic in Appendix \ref{app:implementation_details}, this explains why GAM can preserve efficiency despite relying on LLM-based semantic boundary discrimination.

\subsection{Long-Term Scaling Complexity}
\label{app:long_term_scaling}

In addition to the retrieval-size sensitivity analysis, we analyze how the graph grows under progressive session injection. The key question is whether increasing long-term memory size causes inference to scale poorly. We therefore incrementally inject up to 27 long-horizon sessions and track structural growth together with inference cost.

\begin{table*}[!htbp]
\caption{Complexity evolution of the dual-layer graph memory under progressive session injection.}
\small
\centering
\begin{tabular}{lcccccc}
\toprule
\textbf{Sessions} & \textbf{Event Nodes} & \textbf{Topic Nodes} & \textbf{Edges} & \textbf{Avg. Tokens} & \textbf{Avg. Latency (s)} & \textbf{Avg. F1} \\
\midrule
3 & 81 & 8 & 371 & 1604 & 3.49 & 0.65 \\
15 & 363 & 43 & 1791 & 1864 & 4.16 & 0.57 \\
27 & 657 & 84 & 3307 & 1836 & 4.71 & 0.48 \\
\bottomrule
\end{tabular}
\label{tab:long_term_scaling}
\end{table*}

Although the total number of graph edges increases substantially, the inference token count and latency remain comparatively stable because retrieval first prunes the search space through the topic layer before drilling down into episodic evidence. This supports the claim that GAM avoids uncontrolled growth in query-time overhead even as the long-term memory structure expands.

\subsection{Robustness Analysis on Topic Segmentation Noise}
\label{app:noise}

\subsubsection{Experimental Setup}
To rigorously evaluate the robustness of our GAM framework against errors in the topic boundary detection module $\mathcal{M}_{\theta}$, we designed a composite noise injection mechanism that simulates three distinct types of segmentation failures. These include Miss errors which correspond to false negatives, Shift errors resulting in temporal misalignment, and Extra errors serving as false positives. Specifically, given a noise level $\eta$ ranging from 0.0 to 0.4, we first iterate through the ground-truth boundaries. With a probability of $\eta$, each boundary is either deleted to simulate a missed topic switch or shifted randomly by up to two turns in either direction to simulate imprecise boundary localization. 

Subsequently, to simulate hallucinated boundaries, we inject extra random cut-points into the dialogue where the number of inserted boundaries is proportional to the original count scaled by $\eta$. This setup creates a challenging and highly perturbed environment that tests the system's ability to recover context despite fragmentation, fusion, and misalignment errors.

\subsubsection{Results and Analysis}

\begin{figure}[t]
    \centering
    \includegraphics[width=1.0\linewidth]{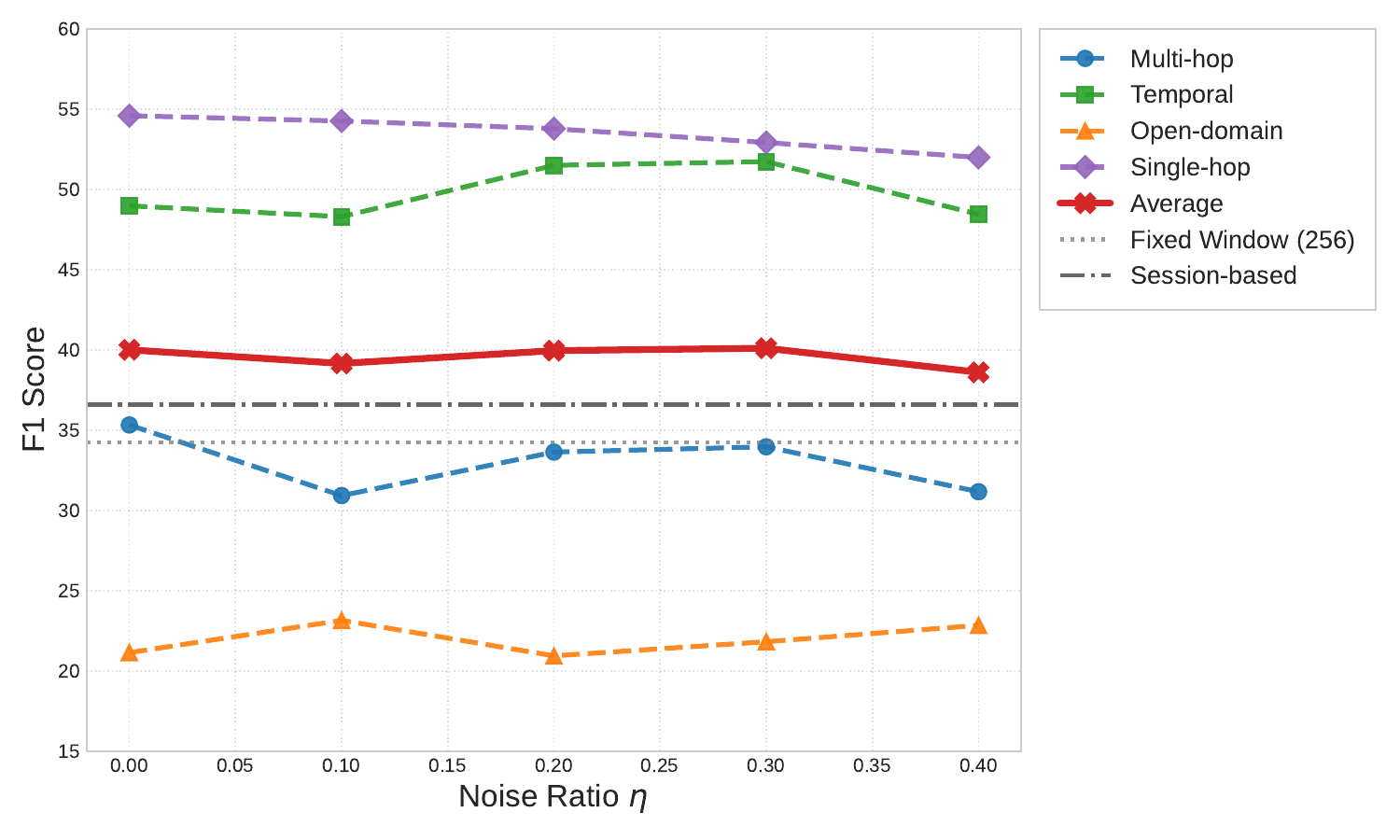}
    \caption{Robustness analysis of the proposed framework under varying topic segmentation noise levels $\eta$. The solid red line represents the Average F1 score, which remains remarkably stable around 40.0, significantly outperforming the Fixed Window (Gray Dotted Line) and Session-based (Black Dash-Dotted Line) baselines even at $\eta=0.4$. Dashed lines indicate performance on specific task categories, highlighting the resilience of the hierarchical graph architecture against boundary errors.}
    \label{fig:noise_robustness}
\end{figure}

As shown in Figure \ref{fig:noise_robustness}, the performance trajectory on the LoCoMo benchmark demonstrates the exceptional stability of our framework against segmentation errors. Despite complex perturbations including boundary deletions and shifts, the Average F1 score remains consistent around 40.0. Even at the most severe noise level where $\eta$ equals 0.4, the model achieves an F1 score of 38.60, which significantly outperforms both the Fixed Window baseline score of 34.23 and the Session-based baseline score of 36.59. These results confirm that the efficacy of our method is not contingent on perfect segmentation, as the underlying Hierarchical Graph Architecture provides a robust scaffolding that organizes information effectively even when structural boundaries are imprecise.

A granular analysis reveals divergent impacts across task categories. Intriguingly, Temporal tasks showed improved performance under moderate noise conditions, peaking at 51.72 when $\eta$ is set to 0.3. This suggests that the random insertions of extra boundaries create an over-segmentation effect, producing finer-grained memory nodes that facilitate precise temporal grounding. In contrast, Single-hop retrieval exhibited a graceful degradation from 54.58 to 51.98, as the shifting or deletion of boundaries can disrupt the local coherence required for keyword matching. However, this decline remains minimal at less than 5\%, demonstrating that our Graph-Guided Multi-Factor Retrieval strategy effectively compensates for structural misalignments by utilizing the semantic connectivity of the graph to bridge gaps created by segmentation errors.
\section{Hyperparameter Analysis}

\subsection{Modulation Factors Sensitivity Analysis}
\label{app:sensitivity}

\begin{figure*}[t]
    \centering
    \includegraphics[width=\textwidth]{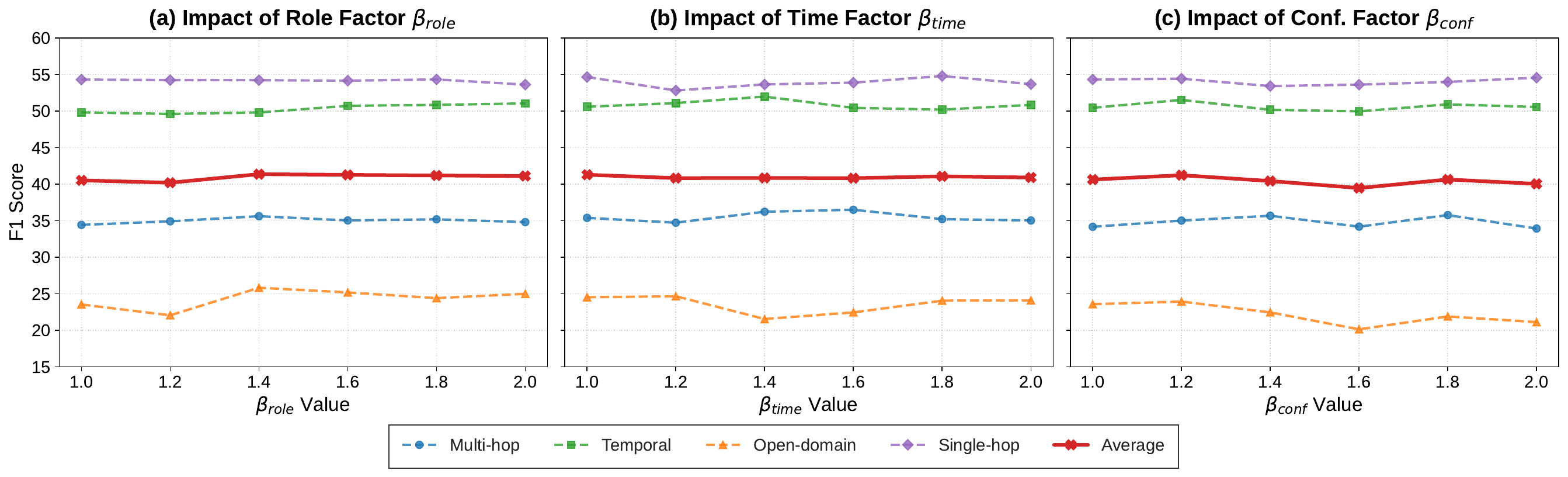}
    \caption{Parameter sensitivity analysis on the LoCoMo dataset using the Qwen 2.5-7B backbone. The three subplots illustrate the impact of varying the (a) Role Factor $\beta_{role}$, (b) Temporal Factor $\beta_{time}$, and (c) Confidence Factor $\beta_{conf}$ on retrieval performance across different task categories. The solid red line represents the Average F1 Score. The relatively flat curves indicate the system's robustness to hyperparameter variations.}
    \label{fig:sensitivity_analysis}
\end{figure*}

To evaluate the robustness of our Graph-Guided Multi-Factor Retrieval mechanism, we conducted a sensitivity analysis on the modulation factors $\beta_{role}$, $\beta_{time}$, and $\beta_{conf}$. We varied each factor from 1.0 to 2.0 while keeping others constant. The results are visualized in Figure \ref{fig:sensitivity_analysis}.

\paragraph{Impact of Role Factor ($\beta_{role}$).}
As illustrated in Figure \ref{fig:sensitivity_analysis}(a), the performance generally improves as the role factor increases, peaking at approximately $\beta_{role}=1.4$. This trend is particularly evident in the Multi-hop task (blue dashed line), confirming that filtering memory candidates based on speaker identity helps the model connect disparate pieces of information associated with specific characters. However, setting the value too high ($\beta > 1.6$) leads to a slight decline, likely because excessive filtering might exclude relevant context mentioned by other speakers in the same scene.

\paragraph{Impact of Temporal Factor ($\beta_{time}$).}
Figure \ref{fig:sensitivity_analysis}(b) demonstrates a distinct trade-off. The Temporal reasoning task (green dashed line) benefits significantly from a higher boosting factor, reaching optimal performance around $\beta_{time}=1.4$. Conversely, the Open-domain task (orange dashed line) experiences a dip at this value. This suggests that while strong temporal modulation sharpens the focus for time-sensitive queries, it may occasionally narrow the retrieval scope for broader, non-temporal questions. We selected 1.4 as the default setting to maximize temporal precision while maintaining a high Average F1 score.

\paragraph{Impact of Confidence Factor ($\beta_{conf}$).}
As shown in Figure \ref{fig:sensitivity_analysis}(c), the system exhibits high stability with respect to the intrinsic confidence factor. The curves remain relatively flat across the tested range, with a slight peak observed between 1.2 and 1.4. This indicates that while prioritizing high-quality memories (those passing the self-test) contributes to overall reliability, the underlying topological structure of the graph plays a more dominant role in retrieval success than the precise tuning of this weight.

Overall, the sensitivity analysis confirms that our multi-factor retrieval mechanism is robust. The standard deviation of the Average F1 score across all tested parameter ranges remains low, implying that the framework's effectiveness derives primarily from its architectural design rather than over-optimized hyperparameters.

\subsection{Retrieval Size Sensitivity Analysis}
\label{app:retrieval_size}

The retrieval size $k$ serves as a critical hyperparameter determining the volume of recalled context. To investigate its impact, we conducted a sensitivity analysis on the LoCoMo dataset, varying $k$ from 5 to 40. 

\begin{figure}[h]
    \centering
    \includegraphics[width=0.8\linewidth]{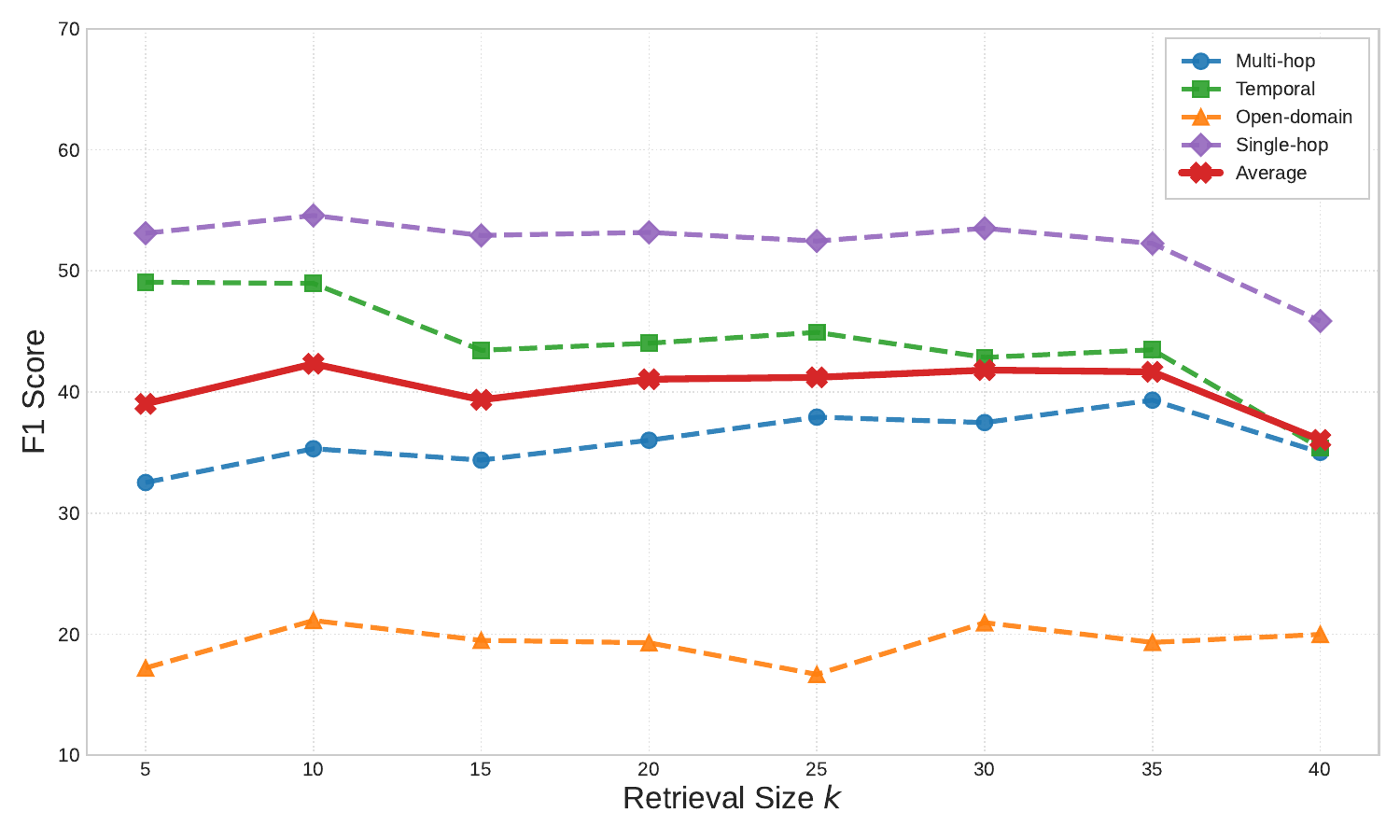} 
    \caption{Impact of retrieval size $k$ on model performance across different task categories. We select $k=10$ as the optimal setting to balance reasoning accuracy with computational efficiency.}
    \label{fig:retrieval_k}
\end{figure}

As shown in Figure \ref{fig:retrieval_k}, the results demonstrate that performance rapidly improves, reaching a peak at $k=10$, but subsequently stagnates or degrades as $k$ increases to 40. This decline suggests that an overly large retrieval set introduces irrelevant noise, which disrupts the model's reasoning. Notably, Temporal tasks are particularly sensitive to this interference, requiring focused windows for chronological precision, whereas Multi-hop tasks remain robust across larger contexts. 
To balance performance and efficiency, we select $k=10$ as the optimal setting, achieving peak accuracy with minimal token consumption compared to larger $k$.

\section{Prompt Engineering Details}
\label{sec:appendix_prompts}

\vspace{-5pt}

To ensure the reproducibility of our experiments, we provide the exact instruction prompts used in the key modules of our framework. All prompts are designed to output structured JSON data to facilitate downstream programmatic processing.

\subsection{Semantic Relation \& Edge Weighting Prompt}
\label{app:prompt_relation}

\vspace{-5pt}

\begin{table}[h!]
    \centering
    \begin{tcolorbox}[
        colback=blue!5, colframe=blue!50,
        title=\textbf{Semantic Relation \& Edge Weighting Prompt},
        boxrule=0.8pt, arc=2pt,
        left=6pt, right=6pt, top=6pt, bottom=6pt
    ]
    \small \ttfamily
    Determine the relationship between the following two memories.
    \vspace{0.5em}
    
    Memory 1: \{memory\_1\_content\}\\
    Memory 2: \{memory\_2\_content\}
    \vspace{0.5em}
    
    Possible relations:
    \begin{itemize}[leftmargin=*, label=-]
        \setlength\itemsep{0em}
        \item "support": one memory supports the other (asymmetric)
        \item "contradict": one memory contradicts the other (asymmetric) 
        \item "coreference": same entity or event (symmetric)
        \item "causal": one memory leads to the other (asymmetric)
        \item "semantic": similar meaning (symmetric)
        \item "unrelated": no meaningful relation
    \end{itemize}
    \vspace{0.5em}
    
    Return JSON strictly in the format: \{\{"relation": "...", "confidence": 0-1\}\}
    \end{tcolorbox}
    \label{tab:prompt_relation}
    \vspace{-10pt}
\end{table}

To construct the global Topic Associative Network in \S\ref{sec:consolidation}, we use this prompt to judge the edge type and weight between nodes. The confidence score (0-1) output by the LLM is directly utilized as the edge weight, allowing the retrieval mechanism to prioritize high-confidence associations.

\subsection{Semantic Node Summarization Prompt}
\label{app:prompt_summary}

During the Semantic Consolidation State (\S\ref{sec:consolidation}), we generate the dual-granularity representation for new topic nodes. This prompt generates the $c_{\text{sum}}$ attribute. We specifically ask for "keywords" and a "concise summary" separately to support both lexical matching and semantic embedding retrieval.

\begin{table}[h!]
    \centering
    \begin{tcolorbox}[
        colback=blue!5, colframe=blue!50,
        title=\textbf{Semantic Node Summarization Prompt},
        boxrule=0.8pt, arc=2pt,
        left=6pt, right=6pt, top=6pt, bottom=6pt
    ]
    \small \ttfamily
    Generate a structured analysis of the following content by:
    \begin{enumerate}[leftmargin=*, label=\arabic*.]
        \setlength\itemsep{0em}
        \item Identifying the most salient keywords and core themes (focus on nouns, verbs, and key concepts).
        \item Writing a concise summary (one to two sentences). Don't be redundant, summarize everything in the fewest words possible.
    \end{enumerate}
    
    Format the response as a JSON object:\\
    \{\\
    \hspace*{1em} "keywords": [\\
    \hspace*{2em} // several specific, distinct keywords...\\
    \hspace*{2em} // Order from most to least important...\\
    \hspace*{1em} ],\\
    \hspace*{1em} "summary": \\
    \hspace*{2em} // a concise one to two sentence summary...\\
    \}
    \vspace{0.5em}
    
    Content for analysis:\\
    \{content\}
    \end{tcolorbox}
    \label{tab:prompt_summary}
    \vspace{-10pt}
\end{table}

\vspace{-5pt}

\subsection{Topic Boundary Detection Prompt}
\label{app:prompt_boundary}

This prompt corresponds to the discriminator function $f_{\theta}$ described in \S\ref{sec:consolidation}. 
We employ a sliding window strategy to provide context. The prompt explicitly instructs the model to return indices in a JSON array to ensure robust parsing by the state machine.

\begin{table}[h!]
    \centering
    \begin{tcolorbox}[
        colback=blue!5, colframe=blue!50,
        title=\textbf{Topic Boundary Detection Prompt},
        boxrule=0.8pt, arc=2pt,
        left=6pt, right=6pt, top=6pt, bottom=6pt
    ]
    \small \ttfamily
    Please analyze the following conversation and identify where topic changes occur. 
    A topic change happens when the conversation shifts to a completely different subject or theme.
    \vspace{0.5em}
    
    Conversation:\\
    \{dialogue\_text\}
    \vspace{0.5em}
    
    Identify the indices (positions) where topic boundaries occur. A boundary at index i means that the topic changes between utterance i and utterance i+1.
    Return only the boundary indices as a JSON array.
    \vspace{0.5em}
    
    Return your response as a JSON object with exactly this structure:\\
    \{\{ "boundaries": [array\_of\_indices] \}\}
    \vspace{0.5em}
    
    Example: \{\{ "boundaries": [2, 6] \}\}
    \end{tcolorbox}
    \label{tab:prompt_boundary}
\end{table}


\begin{thebibliography}{34}
\providecommand{\natexlab}[1]{#1}

\bibitem[{Anokhin et~al.(2024)Anokhin, Semenov, Sorokin, Evseev, Kravchenko,
  Burtsev, and Burnaev}]{anokhin2024arigraph}
Petr Anokhin, Nikita Semenov, Artyom Sorokin, Dmitry Evseev, Andrey Kravchenko,
  Mikhail Burtsev, and Evgeny Burnaev. 2024.
\newblock Arigraph: Learning knowledge graph world models with episodic memory
  for llm agents.
\newblock \emph{arXiv preprint arXiv:2407.04363}.

\bibitem[{Anwar et~al.(2025)Anwar, Welsh, Biswas, Pouya, and
  Chang}]{anwar2025remembr}
Abrar Anwar, John Welsh, Joydeep Biswas, Soha Pouya, and Yan Chang. 2025.
\newblock Remembr: Building and reasoning over long-horizon spatio-temporal
  memory for robot navigation.
\newblock In \emph{2025 IEEE International Conference on Robotics and
  Automation (ICRA)}, pages 2838--2845. IEEE.

\bibitem[{Chang et~al.(2025)Chang, Tang, Wulf, Nyasulu, Wolf, Fernandez-Ruiz,
  and Oliva}]{chang2025sleep}
Hongyu Chang, Wenbo Tang, Annabella~M Wulf, Thokozile Nyasulu, Madison~E Wolf,
  Antonio Fernandez-Ruiz, and Azahara Oliva. 2025.
\newblock Sleep microstructure organizes memory replay.
\newblock \emph{Nature}, 637(8048):1161--1169.

\bibitem[{Chen et~al.(2025)Chen, Niu, Li, Liu, Zheng, Tang, Li, Xiong, and
  Li}]{chen2025halumem}
Ding Chen, Simin Niu, Kehang Li, Peng Liu, Xiangping Zheng, Bo~Tang, Xinchi Li,
  Feiyu Xiong, and Zhiyu Li. 2025.
\newblock Halumem: Evaluating hallucinations in memory systems of agents.
\newblock \emph{arXiv preprint arXiv:2511.03506}.

\bibitem[{Chhikara et~al.(2025)Chhikara, Khant, Aryan, Singh, and
  Yadav}]{chhikara2025mem0}
Prateek Chhikara, Dev Khant, Saket Aryan, Taranjeet Singh, and Deshraj Yadav.
  2025.
\newblock Mem0: Building production-ready ai agents with scalable long-term
  memory.
\newblock \emph{arXiv preprint arXiv:2504.19413}.

\bibitem[{Didolkar et~al.(2024)Didolkar, Goyal, Ke, Guo, Valko, Lillicrap,
  Jimenez~Rezende, Bengio, Mozer, and Arora}]{didolkar2024metacognitive}
Aniket Didolkar, Anirudh Goyal, Nan~Rosemary Ke, Siyuan Guo, Michal Valko,
  Timothy Lillicrap, Danilo Jimenez~Rezende, Yoshua Bengio, Michael~C Mozer,
  and Sanjeev Arora. 2024.
\newblock Metacognitive capabilities of llms: An exploration in mathematical
  problem solving.
\newblock \emph{Advances in Neural Information Processing Systems},
  37:19783--19812.

\bibitem[{Du et~al.(2025)Du, Tian, Ronanki, Rongali, Bodapati, Galstyan, Wells,
  Schwartz, Huerta, and Peng}]{du2025context}
Yufeng Du, Minyang Tian, Srikanth Ronanki, Subendhu Rongali, Sravan Bodapati,
  Aram Galstyan, Azton Wells, Roy Schwartz, Eliu~A Huerta, and Hao Peng. 2025.
\newblock Context length alone hurts llm performance despite perfect retrieval.
\newblock \emph{arXiv preprint arXiv:2510.05381}.

\bibitem[{Edge et~al.(2024)Edge, Trinh, Cheng, Bradley, Chao, Mody, Truitt,
  Metropolitansky, Ness, and Larson}]{edge2024local}
Darren Edge, Ha~Trinh, Newman Cheng, Joshua Bradley, Alex Chao, Apurva Mody,
  Steven Truitt, Dasha Metropolitansky, Robert~Osazuwa Ness, and Jonathan
  Larson. 2024.
\newblock From local to global: A graph rag approach to query-focused
  summarization.
\newblock \emph{arXiv preprint arXiv:2404.16130}.

\bibitem[{Guo et~al.(2024)Guo, Xia, Yu, Ao, and Huang}]{guo2024lightrag}
Zirui Guo, Lianghao Xia, Yanhua Yu, Tu~Ao, and Chao Huang. 2024.
\newblock Lightrag: Simple and fast retrieval-augmented generation.
\newblock \emph{arXiv preprint arXiv:2410.05779}.

\bibitem[{Hu et~al.(2026)Hu, Liu, Yue, Zhang, Liu, Zhu, Lin, Guo, Dou, Xi, Jin,
  Tan, Yin, Liu, Zhang, Sun, Zhu, Sun, Peng, Cheng, Fan, Guo, Yu, Zhou, Hu,
  Huo, Wang, Niu, Wang, Yin, Hu, Liao, Li, Wang, Zhou, Liu, Cheng, Zhang, Gui,
  Pan, Zhang, Torr, Dou, Wen, Huang, Jiang, and Yan}]{hu2026memoryageaiagents}
Yuyang Hu, Shichun Liu, Yanwei Yue, Guibin Zhang, Boyang Liu, Fangyi Zhu,
  Jiahang Lin, Honglin Guo, Shihan Dou, Zhiheng Xi, Senjie Jin, Jiejun Tan,
  Yanbin Yin, Jiongnan Liu, Zeyu Zhang, Zhongxiang Sun, Yutao Zhu, Hao Sun,
  Boci Peng, and 28 others. 2026.
\newblock \href {https://arxiv.org/abs/2512.13564} {Memory in the age of ai
  agents}.
\newblock \emph{Preprint}, arXiv:2512.13564.

\bibitem[{Huang et~al.(2026)Huang, Zhang, Liang, Bei, Chen, Feng, Pan, Tan,
  Wang, Wei, Wu, Xu, Yang, Yang, Yang, Yeh, Zhang, Zhang, Zhu, Zou, Zhao, Wang,
  Xu, Ke, Hui, Li, Wu, He, Wang, Xu, Huang, Tan, Heinecke, Wang, Xiong,
  Metwally, Yan, Lee, Zeng, Xia, Wei, Payani, Wang, Ma, Wang, Wang, Zhang,
  Wang, Zhang, You, Tong, Luo, Liu, Sun, Wang, McAuley, Zou, Han, Yu, and
  Shu}]{huang2026rethinkingmemorymechanismsfoundation}
Wei-Chieh Huang, Weizhi Zhang, Yueqing Liang, Yuanchen Bei, Yankai Chen, Tao
  Feng, Xinyu Pan, Zhen Tan, Yu~Wang, Tianxin Wei, Shanglin Wu, Ruiyao Xu,
  Liangwei Yang, Rui Yang, Wooseong Yang, Chin-Yuan Yeh, Hanrong Zhang, Haozhen
  Zhang, Siqi Zhu, and 41 others. 2026.
\newblock \href {https://arxiv.org/abs/2602.06052} {Rethinking memory
  mechanisms of foundation agents in the second half: A survey}.
\newblock \emph{Preprint}, arXiv:2602.06052.

\bibitem[{Jia et~al.(2025{\natexlab{a}})Jia, Zhang, M\'{e}ndez, and
  Omran}]{jia2025structrag}
Runsong Jia, Bowen Zhang, Sergio Jos\'{e}~Rodr\'{\i}guez M\'{e}ndez, and
  Pouya~G. Omran. 2025{\natexlab{a}}.
\newblock \href {https://doi.org/10.1145/3701716.3717819} {Structrag:
  Structure-aware rag framework with scholarly knowledge graph for diverse
  question answering}.
\newblock In \emph{Companion Proceedings of the ACM on Web Conference 2025},
  page 2567–2573. Association for Computing Machinery.

\bibitem[{Jia et~al.(2025{\natexlab{b}})Jia, Liu, Li, Chen, and
  Liu}]{jia2025evaluating}
Zixi Jia, Qinghua Liu, Hexiao Li, Yuyan Chen, and Jiqiang Liu.
  2025{\natexlab{b}}.
\newblock Evaluating the long-term memory of large language models.
\newblock In \emph{Findings of the Association for Computational Linguistics:
  ACL 2025}, pages 19759--19777.

\bibitem[{Kang et~al.(2025)Kang, Ji, Zhao, and Bai}]{kang2025memory}
Jiazheng Kang, Mingming Ji, Zhe Zhao, and Ting Bai. 2025.
\newblock Memory os of ai agent.
\newblock \emph{arXiv preprint arXiv:2506.06326}.

\bibitem[{Kim et~al.(2024)Kim, Chay, Hwang, Kyung, Chung, Cho, Jo, and
  Choi}]{kim2024dialsim}
Jiho Kim, Woosog Chay, Hyeonji Hwang, Daeun Kyung, Hyunseung Chung, Eunbyeol
  Cho, Yohan Jo, and Edward Choi. 2024.
\newblock Dialsim: A real-time simulator for evaluating long-term multi-party
  dialogue understanding of conversation systems.
\newblock \emph{arXiv preprint arXiv:2406.13144}.

\bibitem[{Lee et~al.(2024)Lee, Chen, Furuta, Canny, and Fischer}]{lee2024human}
Kuang-Huei Lee, Xinyun Chen, Hiroki Furuta, John Canny, and Ian Fischer. 2024.
\newblock A human-inspired reading agent with gist memory of very long
  contexts.
\newblock \emph{arXiv preprint arXiv:2402.09727}.

\bibitem[{Li et~al.(2025)Li, Zhang, Yang, Huang, Wu, Luo, Bei, Zou, Luo, Zhao
  et~al.}]{li2025towards}
Yangning Li, Weizhi Zhang, Yuyao Yang, Wei-Chieh Huang, Yaozu Wu, Junyu Luo,
  Yuanchen Bei, Henry~Peng Zou, Xiao Luo, Yusheng Zhao, and 1 others. 2025.
\newblock Towards agentic rag with deep reasoning: A survey of rag-reasoning
  systems in llms.
\newblock \emph{arXiv preprint arXiv:2507.09477}, 2.

\bibitem[{Ma et~al.(2025)Ma, Chen, Wu, Khan, and Wang}]{ma2025large}
Chuangtao Ma, Yongrui Chen, Tianxing Wu, Arijit Khan, and Haofen Wang. 2025.
\newblock Large language models meet knowledge graphs for question answering:
  Synthesis and opportunities.
\newblock \emph{arXiv preprint arXiv:2505.20099}.

\bibitem[{Maharana et~al.(2024)Maharana, Lee, Tulyakov, Bansal, Barbieri, and
  Fang}]{maharana2024evaluating}
Adyasha Maharana, Dong-Ho Lee, Sergey Tulyakov, Mohit Bansal, Francesco
  Barbieri, and Yuwei Fang. 2024.
\newblock Evaluating very long-term conversational memory of llm agents.
\newblock In \emph{Proceedings of the 62nd Annual Meeting of the Association
  for Computational Linguistics (Volume 1: Long Papers)}, pages 13851--13870.

\bibitem[{Packer et~al.(2024)Packer, Wooders, Lin, Fang, Patil, Stoica, and
  Gonzalez}]{packer2024memgptllmsoperatingsystems}
Charles Packer, Sarah Wooders, Kevin Lin, Vivian Fang, Shishir~G. Patil, Ion
  Stoica, and Joseph~E. Gonzalez. 2024.
\newblock \href {https://arxiv.org/abs/2310.08560} {Memgpt: Towards llms as
  operating systems}.
\newblock \emph{Preprint}, arXiv:2310.08560.

\bibitem[{Park et~al.(2023)Park, O'Brien, Cai, Morris, Liang, and
  Bernstein}]{park2023generative}
Joon~Sung Park, Joseph O'Brien, Carrie~Jun Cai, Meredith~Ringel Morris, Percy
  Liang, and Michael~S Bernstein. 2023.
\newblock Generative agents: Interactive simulacra of human behavior.
\newblock In \emph{Proceedings of the 36th annual acm symposium on user
  interface software and technology}, pages 1--22.

\bibitem[{Rasmussen et~al.(2025)Rasmussen, Paliychuk, Beauvais, Ryan, and
  Chalef}]{rasmussen2025zep}
Preston Rasmussen, Pavlo Paliychuk, Travis Beauvais, Jack Ryan, and Daniel
  Chalef. 2025.
\newblock Zep: a temporal knowledge graph architecture for agent memory.
\newblock \emph{arXiv preprint arXiv:2501.13956}.

\bibitem[{Su et~al.(2026)Su, Guo, Hou, Bai, Li, Zhang, Yin, Lin, Jin, Guo, and
  Cheng}]{su2026dialoguetimetemporalsemantic}
Miao Su, Yucan Guo, Zhongni Hou, Long Bai, Zixuan Li, Yufei Zhang, Guojun Yin,
  Wei Lin, Xiaolong Jin, Jiafeng Guo, and Xueqi Cheng. 2026.
\newblock \href {https://arxiv.org/abs/2601.07468} {Beyond dialogue time:
  Temporal semantic memory for personalized llm agents}.
\newblock \emph{Preprint}, arXiv:2601.07468.

\bibitem[{Tan et~al.(2025)Tan, Yan, Hsu, Han, Wang, Le, Song, Chen, Palangi,
  Lee, Iyer, Chen, Liu, Lee, and Pfister}]{tan2025prospect}
Zhen Tan, Jun Yan, I-Hung Hsu, Rujun Han, Zifeng Wang, Long Le, Yiwen Song,
  Yanfei Chen, Hamid Palangi, George Lee, Anand~Rajan Iyer, Tianlong Chen, Huan
  Liu, Chen-Yu Lee, and Tomas Pfister. 2025.
\newblock \href {https://aclanthology.org/2025.acl-long.413/} {In prospect and
  retrospect: Reflective memory management for long-term personalized dialogue
  agents}.
\newblock In \emph{Proceedings of the 63rd Annual Meeting of the Association
  for Computational Linguistics (Volume 1: Long Papers)}, pages 8416--8439.
  Association for Computational Linguistics.

\bibitem[{Xu et~al.(2025)Xu, Liang, Mei, Gao, Tan, and Zhang}]{xu2025mem}
Wujiang Xu, Zujie Liang, Kai Mei, Hang Gao, Juntao Tan, and Yongfeng Zhang.
  2025.
\newblock A-mem: Agentic memory for llm agents.
\newblock \emph{arXiv preprint arXiv:2502.12110}.

\bibitem[{Yang et~al.(2025{\natexlab{a}})Yang, Tang, Shen, Guo, Wang, Cao, and
  Zhang}]{yang2025continual}
Enneng Yang, Anke Tang, Li~Shen, Guibing Guo, Xingwei Wang, Xiaochun Cao, and
  Jie Zhang. 2025{\natexlab{a}}.
\newblock Continual model merging without data: Dual projections for balancing
  stability and plasticity.
\newblock In \emph{The Thirty-ninth Annual Conference on Neural Information
  Processing Systems}.

\bibitem[{Yang et~al.(2025{\natexlab{b}})Yang, Hu, Li, Wang, Yu, Xu, Zhu, and
  Yao}]{yang2025drunkagent}
Shiyi Yang, Zhibo Hu, Xinshu Li, Chen Wang, Tong Yu, Xiwei Xu, Liming Zhu, and
  Lina Yao. 2025{\natexlab{b}}.
\newblock Drunkagent: Stealthy memory corruption in llm-powered recommender
  agents.
\newblock \emph{arXiv preprint arXiv:2503.23804}.

\bibitem[{Zhang et~al.(2025{\natexlab{a}})Zhang, Fu, Wan, Yu, Wang, and
  Yan}]{zhang2025g}
Guibin Zhang, Muxin Fu, Guancheng Wan, Miao Yu, Kun Wang, and Shuicheng Yan.
  2025{\natexlab{a}}.
\newblock G-memory: Tracing hierarchical memory for multi-agent systems.
\newblock \emph{arXiv preprint arXiv:2506.07398}.

\bibitem[{Zhang et~al.(2026{\natexlab{a}})Zhang, Fan, Zou, Chen, Wang, Zhou,
  Li, Huang, Yao, Zheng, Liu, Li, and
  Yu}]{zhang2026evoskillsselfevolvingagentskills}
Hanrong Zhang, Shicheng Fan, Henry~Peng Zou, Yankai Chen, Zhenting Wang, Jiayu
  Zhou, Chengze Li, Wei-Chieh Huang, Yifei Yao, Kening Zheng, Xue Liu, Xiaoxiao
  Li, and Philip~S. Yu. 2026{\natexlab{a}}.
\newblock \href {https://arxiv.org/abs/2604.01687} {Evoskills: Self-evolving
  agent skills via co-evolutionary verification}.
\newblock \emph{Preprint}, arXiv:2604.01687.

\bibitem[{Zhang et~al.(2025{\natexlab{b}})Zhang, Li, Bei, Luo, Wan, Yang, Xie,
  Yang, Huang, Miao, Zou, Luo, Zhao, Chen, Chan, Zhou, Zhang, Zhang, Shang,
  Zhang, Song, King, and Yu}]{zhang2025websearchagenticdeep}
Weizhi Zhang, Yangning Li, Yuanchen Bei, Junyu Luo, Guancheng Wan, Liangwei
  Yang, Chenxuan Xie, Yuyao Yang, Wei-Chieh Huang, Chunyu Miao, Henry~Peng Zou,
  Xiao Luo, Yusheng Zhao, Yankai Chen, Chunkit Chan, Peilin Zhou, Xinyang
  Zhang, Chenwei Zhang, Jingbo Shang, and 4 others. 2025{\natexlab{b}}.
\newblock \href {https://arxiv.org/abs/2506.18959} {From web search towards
  agentic deep research: Incentivizing search with reasoning agents}.
\newblock \emph{Preprint}, arXiv:2506.18959.

\bibitem[{Zhang et~al.(2026{\natexlab{b}})Zhang, Zhang, Yu, Nie, Wu, Yue, Liu,
  and Li}]{zhang2026expseek}
Wenyuan Zhang, Xinghua Zhang, Haiyang Yu, Shuaiyi Nie, Bingli Wu, Juwei Yue,
  Tingwen Liu, and Yongbin Li. 2026{\natexlab{b}}.
\newblock \href {https://arxiv.org/abs/2601.08605} {Expseek: Self-triggered
  experience seeking for web agents}.
\newblock \emph{Preprint}, arXiv:2601.08605.

\bibitem[{Zhong et~al.(2024)Zhong, Guo, Gao, Ye, and
  Wang}]{zhong2024memorybank}
Wanjun Zhong, Lianghong Guo, Qiqi Gao, He~Ye, and Yanlin Wang. 2024.
\newblock Memorybank: Enhancing large language models with long-term memory.
\newblock In \emph{Proceedings of the AAAI Conference on Artificial
  Intelligence}, volume~38, pages 19724--19731.

\bibitem[{Zou et~al.(2025)Zou, Huang, Wu, Chen, Miao, Nguyen, Zhou, Zhang,
  Fang, He, Li, Li, Jiang, Liu, and
  Yu}]{zou2025llmbasedhumanagentcollaborationinteraction}
Henry~Peng Zou, Wei-Chieh Huang, Yaozu Wu, Yankai Chen, Chunyu Miao, Hoang
  Nguyen, Yue Zhou, Weizhi Zhang, Liancheng Fang, Langzhou He, Yangning Li,
  Dongyuan Li, Renhe Jiang, Xue Liu, and Philip~S. Yu. 2025.
\newblock \href {https://arxiv.org/abs/2505.00753} {Llm-based human-agent
  collaboration and interaction systems: A survey}.
\newblock \emph{Preprint}, arXiv:2505.00753.

\bibitem[{Zou et~al.(2026)Zou, Miao, Huang, Chen, Zhou, Zhang, Wu, Fang, Gu,
  Zhang, Zheng, Wang, Nian, Li, Fan, He, Zhang, Liu, and
  Yu}]{zou2026userschangemindevaluating}
Henry~Peng Zou, Chunyu Miao, Wei-Chieh Huang, Yankai Chen, Yue Zhou, Hanrong
  Zhang, Yaozu Wu, Liancheng Fang, Zhengyao Gu, Zhen Zhang, Kening Zheng,
  Fangxin Wang, Yi~Nian, Shanghao Li, Wenzhe Fan, Langzhou He, Weizhi Zhang,
  Xue Liu, and Philip~S. Yu. 2026.
\newblock \href {https://arxiv.org/abs/2604.00892} {When users change their
  mind: Evaluating interruptible agents in long-horizon web navigation}.
\newblock \emph{Preprint}, arXiv:2604.00892.

\end{thebibliography}
\end{document}